\pdfoutput=1
\documentclass[11pt]{article}

\usepackage{EMNLP2023}
\usepackage{times}
\usepackage{latexsym}
\usepackage[figuresright]{rotating}
\usepackage{amsmath}
\usepackage{xcolor}
\usepackage[T1]{fontenc}
\usepackage[utf8]{inputenc}
\usepackage{microtype}
\usepackage{inconsolata}
\usepackage{booktabs}
\usepackage{graphicx}
\usepackage{subcaption}
\usepackage{multirow}
\usepackage{bm}
\usepackage{amsthm,amsmath,amssymb}
\usepackage{color}
\usepackage{colortbl}
\usepackage{tcolorbox}
\usepackage{xcolor}
\usepackage{mathrsfs}
\usepackage{CJKutf8}
\usepackage{hyperref} 
\usepackage{ruby}
\usepackage{xpinyin}
\usepackage[english]{babel} % English as the main language
\babelprovide[import]{hindi}

\newtcbox{\red}[1][]{on line,boxsep=2pt,boxrule=0pt,left=0pt,right=0pt,top=0pt,bottom=0pt,colframe=white,colback=red!30!white,#1}

\newtcbox{\blue}[1][]{on line,boxsep=2pt,boxrule=0pt,left=0pt,right=0pt,top=0pt,bottom=0pt,colframe=white,colback=blue!30!white,#1}

\newcommand{\increasebg}[1]{{\cellcolor[HTML]{f4dee0}{#1}}}
\definecolor{mylightred}{HTML}{f4dee0}
\newtcbox{\lightred}[1][]{on line,boxsep=2pt,left=0pt,right=0pt,top=0pt,bottom=0pt,colframe=white,colback=mylightred!100!white,#1}

\title{Cultural Compass: Predicting Transfer Learning Success in Offensive Language Detection with Cultural Features}
\newcommand{\uestc}{$^1$}
\newcommand{\ku}{$^2$}

\author{Li Zhou\uestc \ku, Antonia Karamolegkou\ku, Wenyu Chen\uestc, Daniel Hershcovich\ku \\
{\uestc}University of Electronic Science and Technology of China\\
{\ku}Department of Computer Science, University of Copenhagen \\
\texttt{\small li\_zhou@std.uestc.edu.cn,
antka@di.ku.dk, cwy@uestc.edu.cn, dh@di.ku.dk}
}

\begin{document}
\maketitle
\begin{abstract}
The increasing ubiquity of language technology necessitates a shift towards considering cultural diversity in the machine learning realm, particularly for subjective tasks that rely heavily on cultural nuances, such as Offensive Language Detection (OLD). 
Current understanding underscores that these tasks are substantially influenced by cultural values, however, a notable gap exists in determining if cultural features can accurately predict the success of cross-cultural transfer learning for such subjective tasks. 
Addressing this, our study delves into the intersection of cultural features and transfer learning effectiveness. The findings reveal that cultural value surveys indeed possess a predictive power for cross-cultural transfer learning success in OLD tasks and that it can be further improved using offensive word distance. 
Based on these results, we advocate for the integration of cultural information into datasets. 
Additionally, we recommend leveraging data sources rich in cultural information, such as surveys, to enhance cultural adaptability. 
Our research signifies a step forward in the quest for more inclusive, culturally sensitive language technologies.\footnote{Code is available in \url{https://github.com/lizhou21/cultural-compass}.}

\textbf{\small Warning}: \emph{\small This paper discusses examples of offensive content. The authors do not support the use of offensive language, nor any of the offensive representations quoted below.}

\end{abstract}

\section{Introduction}

Cross-lingual transfer has been successful in various NLP tasks \cite{plank-agic-2018-distant, rahimi-etal-2019-massively, eskander-etal-2020-unsupervised, adelani-etal-2022-masakhaner}
%like named entity recognition~\cite{xie-etal-2018-neural, rahimi-etal-2019-massively, adelani-etal-2022-masakhaner}, entity linking~\cite{rijhwani2019zero, schumacher-etal-2021-cross}, and part-of-speech tagging~\cite{plank-agic-2018-distant, eskander-etal-2020-unsupervised}, 
particularly in scenarios where low-resource languages lack sufficient training data \cite{lauscher-etal-2020-zero}. To ensure successful cross-lingual transfer, leveraging prior knowledge becomes crucial in selecting the most suitable transfer language and dataset. Most research studies predominantly rely on data-dependent and language-dependent features to predict the optimal language to transfer from~\cite{lin-etal-2019-choosing, sun-etal-2021-cross}. However, the same language is often spoken in vastly different cultures, and conversely, similar cultures may speak very different languages. %certain languages are recognized as official languages by multiple countries located in diverse geographical regions.\footnote{\href{https://en.wikipedia.org/wiki/List_of_languages_by_the_number_of_countries_in_which_they_are_recognized_as_an_official_language}{\texttt{List of languages by the number of countries in which they are recognized as an official language.}}}
In subjective tasks such as Offensive Language Detection (OLD), relying solely on measures of linguistic proximity and data-specific features for predicting the optimal transfer language/dataset may be insufficient. It is crucial to consider cultural differences.

% Cultural background has a substantive effect on musical emotion~\cite{10.1093/acprof:oso/9780199230143.003.0027}, pragmatic analysis ~\cite{karoui-etal-2017-exploring, oster2019cross}, writing style~\cite{ma-etal-2022-encbp}, time expressions~\cite{shwartz-2022-good}, and other aspects involving psychology and cognition. 
%We regard such NLP tasks as cultural-laden tasks. 
Cultural sensitivity and adaptation pose a great challenge in NLP tasks \cite{PAPADAKIS2022109031, arango-monnar-etal-2022-resources, hershcovich-etal-2022-challenges, sittar2023classification}. Despite the capabilities of Language Models like ChatGPT, their cultural adaptation to culturally diverse human societies is limited \cite{cao-etal-2023-assessing}. In an increasingly globalized and AI-driven world, this gap needs to be addressed to create and sustain an effective communication environment~\cite{aririguzoh2022communication}.

% \begin{figure}
% \centering
% \includegraphics[width=0.9\linewidth]{FIG/pipeline.png}
% \caption{Approach for selecting high-resource transfer datasets. On the left, exhaustive method is employed to choose the optimal transfer dataset, while on the right, ranking prediction is conducted based on the constructed feature space of the transfer datasets. The predicted ranking results are utilized to select the most promising transfer dataset for the target dataset.}
% % (b) The top 7 languages by the number of countries in which they are recognized as an official language. }
% \label{fig:pipeline}
% \end{figure}

Taking steps towards this direction, we focus on the success of cross-cultural transfer learning in OLD. Offensive language can vary greatly depending on cultural backgrounds.
While most multi-lingual OLD datasets are constructed by filtering a predefined list of offensive words~\cite{zampieri-etal-2019-predicting, sigurbergsson-derczynski-2020-offensive, jeong-etal-2022-kold, deng-etal-2022-cold}, 
% it is important to note that
certain offensive words are culturally specific.
For example, OLD models trained on American cultural contexts may struggle to effectively detect offensive words like ``m*adarchod'' and ``pr*sstitute'' in Indian texts~\cite{ghosh-etal-2021-detecting, santyliang2023nlpositionality}.
Previous studies~\cite{nozza-2021-exposing,litvak-etal-2022-offensive, zhou-etal-2023-cross} have highlighted the challenges of cross-lingual transfer in OLD, emphasizing the significance of accurately predicting the optimal language/dataset for transfer before the implementation. 
% Figure\ref{fig:pipeline} visually illustrates the necessity of transfer prediction.

% The challenges of cross-cultural OLD are shown in Figure~\ref{fig:TL_LLM}, in which we adopt zero-shot transfer learning setting to explore. We see that for the same target dataset, different source datasets have different negative effects. When faced with new low-resource target datasets, it is difficult to select suitable source datasets for transfer learning. 
% Representing culture solely through language is problematic due to factors like colonization and immigration. Some languages are recognized as official languages by multiple countries located in different geographical regions.\footnote{\href{https://en.wikipedia.org/wiki/List_of_languages_by_the_number_of_countries_in_which_they_are_recognized_as_an_official_language}{\texttt{List of languages by the number of countries in which they are recognized as an official language.}}} Therefore, relying solely on measures of linguistic proximity for transfer prediction in cultural-laden tasks might be insufficient.

% To address this problem, 

Given this context, we propose six country-level features that quantify the variations in cultural values across different countries. Additionally, we introduce a %language-level 
feature that measures the %distinctions 
distance %in the expression 
of offensive words between languages. We address three research questions regarding cross-cultural/cross-lingual transfer learning for OLD: 
\begin{description}
    \item[RQ1.] Can we predict transfer learning success?
    \item[RQ2.] Which types of features (cultural or offensive word distance) are most informative?
    \item[RQ3.] Do the same features contribute to transfer learning success in OLD and other NLP tasks?
\end{description}
% (1) Whether it is possible to predict the success of transfer learning before its implementation. (2) Which types of features (cultural or offensive word distance measures) are most informative in determining the transfer learning success. (3) Whether the features that contribute to transfer learning success in OLD are also important for other NLP tasks such as sentiment analysis (SA), dependency parsing (DEP).
Our experiments show the substantial influence of integrating cultural values in predicting language transfer for a culture-loaded task, OLD.
%Our results show the substantial influence of integrating cultural values in predicting language transfer for culture-loaded task OLD. %Additionally, we discover that exploring word distance in specific domains is beneficial for transfer prediction. However, we observed that the majority of current datasets lack adequate cultural information, which poses a hindrance to the advancement of cross-cultural NLP research.
We also find that exploring offensive word distance in specific domains benefits the prediction of the optimal transfer datasets. Notably, we observe that the majority of current datasets lack sufficient cultural information, posing a hindrance to the advancement of cross-cultural NLP research. We adopt \textsc{HarmCheck}~\cite{kirk-etal-2022-handling} in Appendix~\ref{app:harmcheck} for handling and presenting harmful text in research.

\section{Related Work}

\paragraph{OLD methods.} 
There have been many works on offensive language detection.
Some works concentrate more on how to improve OLD systems~\cite{goel-sharma-2022-leveraging, mcgillivray-etal-2022-leveraging, liu-etal-2022-multiple}.
% \citet{goel-sharma-2022-leveraging} incorporate syntactic features by utilizing graph convolutional networks to improve the detecting ability. 
% \citet{mcgillivray-etal-2022-leveraging} claim the meanings of words are constantly evolving and enrich a lexicon-based OLD system with time-sensitive lexical features, which shows indeed language change affects offensive language.
% \citet{liu-etal-2022-multiple} formalize OLD into a multiple-instance learning task to utilize coarse-grained natural labels from online platforms. 
Considering that offensive language online is a worldwide issue and research efforts should not be confined solely to English~\cite{sogaard-2022-ban}, some researchers focus on constructing non-English OLD datasets~\cite{coltekin-2020-corpus, sigurbergsson-derczynski-2020-offensive, mubarak-etal-2021-arabic, deng-etal-2022-cold, jeong-etal-2022-kold}.
Based on these datasets, some works try to apply cross-lingual transfer learning for LOD~\cite{nozza-2021-exposing, litvak-etal-2022-offensive, arango-monnar-etal-2022-resources, ERONEN2022, zhou-etal-2023-cross},
%but performance has so far been significantly less impressive. % That is because OLD is a very cultural-laden task, which is strongly influenced by cultural factors.
but performance has so far not taken into account cultural factors. Understanding the cultural nuances and context surrounding offensive language is crucial for developing effective and accurate models for detecting such language.

%\cite{ERONEN2022} demonstrate the effectiveness of cross-lingual transfer learning for zero-shot abusive language detection, with the selection of an optimal transfer language based on linguistic similarity metrics.

\paragraph{Optimal transfer language prediction.}
A major challenge in cross-lingual learning is choosing the optimal transfer language. 
Language proximity is not always the best criterion, since there are other linguistic properties that could lead to better results such as phonological or syntactic distances~\cite{karamolegkou-stymne-2021-investigation, ERONEN2022}. 
Other factors that can influence transfer language selection, such as lexical overlap, have shown mixed findings. Some studies report a positive correlation with the cross-lingual model performance \cite{wu-dredze-2019-beto, patil-etal-2022-overlap, de-vries-etal-2022-make}, while others do not support this finding \cite{pires-etal-2019-multilingual, tran-bisazza-2019-zero, conneau-etal-2020-emerging}.
To automate the process of selecting transfer languages, there have been attempts using a ranking meta-model that predicts the most optimal languages \cite{lin-etal-2019-choosing, lauscher-etal-2020-zero, srinivasan2021predicting, Doliki2021AnalysingTI, Srinivasan_2022, ahuja-etal-2022-multi, patankar-etal-2022-train}. These mostly rely on data-dependent and language-dependent features, without taking into account cultural background differences during the optimal transfer language prediction process.

% Choosing Transfer Languages for Cross-Lingual Learning
% Predicting Performance for Natural Language Processing Tasks

\paragraph{Cultural features.}
% In Arabic, many offensive tweets have the vocative particle~\cite{mubarak-etal-2021-arabic}.
Recent research has begun to focus on cross-cultural NLP~\cite{hershcovich-etal-2022-challenges}.
Cultural feature augmentation is able to improve the performance of deep learning models on various semantic, syntactic, and psycholinguistic tasks~\cite{ma-etal-2022-encbp}.
\citet{sun-etal-2021-cross} introduce three linguistic features that capture cross-cultural similarities evident in linguistic patterns and quantify distinct aspects of language pragmatics. Building upon these features, they extend the existing research on auxiliary language selection in cross-lingual tasks.
% \citet{ghosh-etal-2021-detecting} introduce a weakly supervised method to robustly detect lexical biases in broader geocultural contexts.
However, \citet{lwowski-etal-2022-measuring} confirms that offensive language models exhibit geographical variations, even when applied to the same language. This suggests that using language as a proxy for considering cultural differences is overly simplistic. \citet{santyliang2023nlpositionality} highlight the impact of researcher positionality, which introduces design biases and subsequently influences the positionality of datasets and models. 
So in this paper, we consider country information for annotators across various OLD datasets and propose more fine-grained cultural features to enhance the prediction accuracy of transfer learning for cultural-loaded tasks.

% \cite{hardalov2022few} Few-shot cross-lingual stance detection with sentiment-based pre-training

% \citet{sun-etal-2021-cross} introduce three linguistic features that capture cross-cultural similarities that manifest in linguistic patterns and quantify distinct aspects of language pragmatics: language context-level, figurative language, and the lexification of emotion concepts.  quantify cross-cultural similarity, focusing on semantic and pragmatic differences across languages.

% Capturing Cultural Differences in Expressions of Intentions~\cite{tomlinson-etal-2014-capturing}:In this contribution we introduce a novel model which captures latent cultural dimensions through an individual’s expressions of intentionality.  We then  show how these latent cultures can be used to create a culturally-sensitive model which provides  enahnced detection of signals of intentionality in tweets.  Finally, we demonstrate how these  models reveal interesting cross-cultural differences in the goals and motivations of individuals from different cultures.

% \begin{figure}
% \includegraphics[width=1\linewidth]{FIG/lang_count.png}
% \caption{The top 7 languages by the number of countries in which they are recognized as an official language. }
% \label{fig:lang_count}
% \end{figure}

\section{How to predict optimal transfer datasets?}
% Problem Formulation
In this section, we formalize the problem as a \textit{transfer dataset ranking} task for cross-cultural/lingual transfer learning. 
Different from cross-lingual transfer learning, our experiments involve considering datasets from the same language but with different cultural backgrounds.

Specifically, we give a task $t$ and provide a related dataset set $\mathcal{D}=\{d_1, d_2, \dots, d_{n}\}$. 
Our objective is to develop a ranking method for the low-resource target dataset $d_i^{\mathrm{tgt}} \in \mathcal{D}$, which ranks the other $n-1$ candidate high re-resource transfer datasets $\mathcal{D}^{\mathrm{tsf}} = \mathcal{D} - d_i^{\mathrm{tgt}} =\{d_1^{\mathrm{tsf}}, d_2^{\mathrm{tsf}}, \dots, d_{n-1}^{\mathrm{tsf}}\}$.
Exhaustive method is the most direct and simple approach, involving training the task model on each candidate transfer dataset and then evaluating it on the target dataset. However, it is time-consuming and resource-wasting.
% and a related dataset set $\mathcal{D}=\{d_1, d_2, …, d_{n}\}$, 
% our objective is to construct a ranking method $\mathcal{F}(d^{\mathrm{tgt}}, \mathcal{D}^{\mathrm{tsf}})$ that generates an optimal rank $\mathrm{R}[\mathcal{D} ^{\mathrm{tsf}}]$ for the remaining high-resource transfer datasets $\mathcal{D} ^{\mathrm{tsf}}=\mathcal{D} -\mathrm{D}^{\mathrm{tgt}}$, with respect to their transferability to the low-resource dataset $\mathrm{D}^{\mathrm{tgt}}\in \mathcal{D}$. Mathematically, we can represent this as:
% \begin{equation}
% \begin{array}{cc}
%     \mathrm{R}\left[ \mathcal{D} ^{\mathrm{tsf}}\right] =\mathcal{F}\left( \mathrm{D}^{\mathrm{tgt}}, \mathcal{D} ^{\mathrm{tsf}}\right), \\

%     \mathrm{s.t.} \; \mathrm{R}\left( \mathrm{D} \right) \ne \mathrm{R}\left( \mathrm{D}' \right), \text{for} \; \forall \;\mathrm{D}, \mathrm{D}' \in \mathcal{D} ^{\mathrm{tsf}},   \\
% \end{array}
% \end{equation}
% where $\mathrm{R}[\mathcal{D} ^{\mathrm{tsf}}]$ represents the rank of the transfer datasets in $\mathcal{D}^{\mathrm{tsf}}$ produced by the ranking method $\mathcal{F}$, the constraint $\mathrm{R}\left( \mathrm{D} \right) \ne \,\,\mathrm{R}\left( \mathrm{D}' \right)$ ensures that each dataset in $\mathcal{D}^{\mathrm{tsf}}$ is assigned a unique rank, maintaining the ordering property of the ranking task.

An effective approach is to utilize the extracted features from the dataset pairs to predict prior knowledge for a cross-cultural task, eliminating the necessity of conducting task-specific experiments.
\[\nu _{d_{i}^{tgt}, d_{j}^{tsf}}=\mathrm{extract}\left( d_{i}^{tgt}, d_{j}^{tsf} \right). \]
These extracted features encompass various aspects, including statistical properties, linguistic information, and domain-specific characteristics of the datasets. 
These features are employed to predict a relative score for each transfer dataset pair.
\[r_{d_{i}^{tgt}, d_{j}^{tsf}}=\mathcal{R}\left( \nu _{d_{i}^{tgt}, d_{j}^{tsf}}; \theta \right), \]
% \begin{equation}
%     \mathrm{R}[ \mathcal{D} ^{\mathrm{tsf}} ] =\mathcal{R} ( \mathcal{V} ( \mathrm{D}^{\mathrm{tgt}}, \mathcal{D} ^{\mathrm{tsf}} ) ), 
% \end{equation}
where $\mathcal{R}(\cdot)$ denotes the ranking model, $\theta$ is the parameters of the ranking model.
% $\mathcal{V}(\cdot)$ presents the feature extractor for each pair of datasets $\left<\mathrm{D}^{\mathrm{tsf}}, \mathrm{D}^{\mathrm{tgt}}\right>$, and 

% So can we successfully predict prior knowledge of a cross-cultural task by training a ranking model that utilizes the extracted features from the datasets?
% Can we predict the success of cross-cultural/cross-lingual transfer learning for offensive language detection a-priori?

% we are presented with a set of diverse candidate transfer datasets $\mathbf{\mathcal{D}} ^{tsf}=\left\{ \mathcal{D} _{1}^{tsf}, \mathcal{D} _{2}^{tsf}, …, \mathcal{D} _{n}^{tsf} \right\} $. Our objective is to rank these datasets according to their transferability to a new low-resource dataset $\mathcal{D} ^{tgt}$.  

% , given a datasets set $\mathcal{D} =\left\{ \mathcal{D} _1, \mathcal{D} _1, …, \mathcal{D} _n \right\}$, we need pre 

\begin{figure*}[htbp]
  \begin{minipage}[ht]{\columnwidth}
    \centering
    \includegraphics[width=\columnwidth]{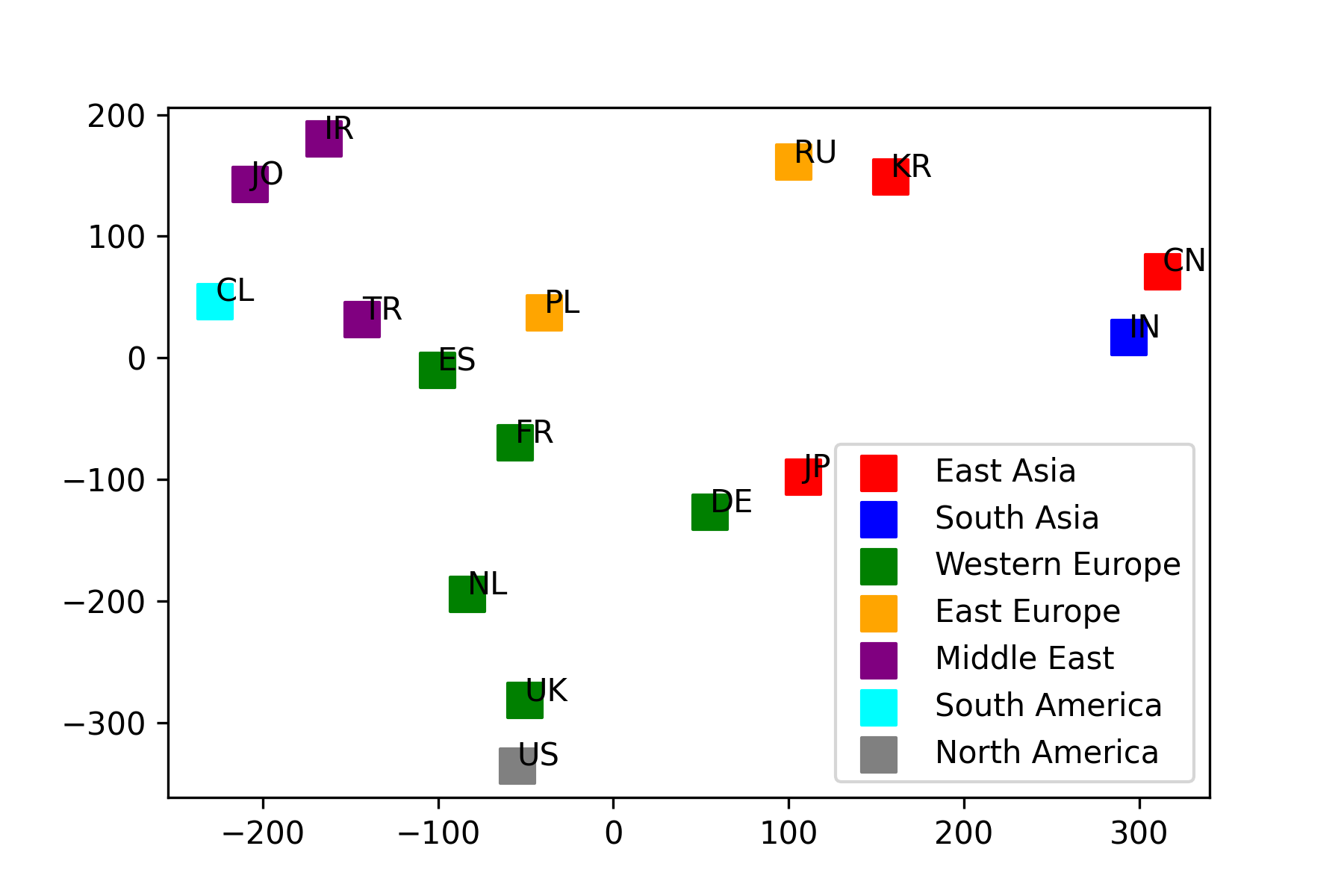}
    \caption{T-SNE Visualization of Cultural Values: Countries represented as data points, with color-coded regions highlighting similarities and differences in cultural dimensions.}
    \label{fig:analysis_cultural}
  \end{minipage}
  \hfill
  \begin{minipage}[ht]{\columnwidth}
    \centering    \includegraphics[width=\columnwidth]{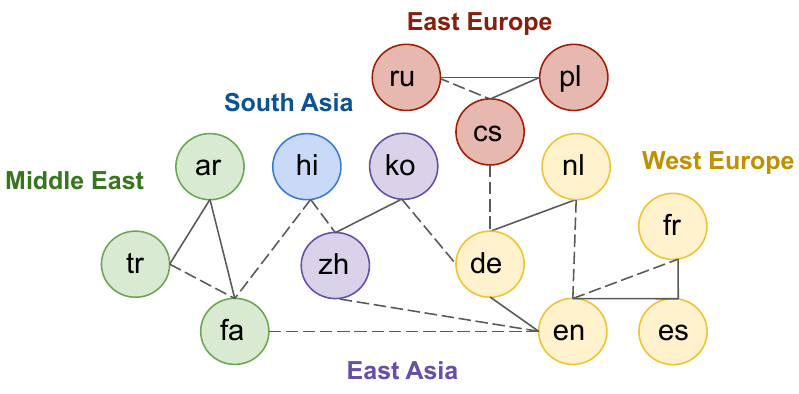}
    \caption{Network based on Offensive Word Distance. Edges connect languages based on their top-2 closest rankings. Solid lines represent mutual ranking, while dashed lines represent exclusive ranking.}
    \label{fig:analysis_off}
  \end{minipage}
\end{figure*}

\section{What features can be used to predict?}
The current features used for transfer learning prediction can be divided into data-dependent and language-dependent categories. Data-dependent features capture statistical characteristics specific to the dataset, while language-dependent features assess language similarities from different perspectives. Appendix~\ref{app:D-L-features} provides further details on these features. However, in cultural-loaded tasks, it is crucial to account for cultural differences. To address this, we propose incorporating six country-level features that quantify variations in cultural values across countries. Furthermore, we introduce a language-level and domain-specific feature that measures the distance between offensive words in language pairs.
% Additionally, we introduce a language-level and domain-specific feature that measures the distance of offensive words between of language pairs.

\subsection{Cultural-Dimension Features}
The country-level features encompass six cultural dimensions that capture cross-cultural variations in values, as per Hofstede's theory~\cite{hofstede1984culture}, across different countries.
% The original model included four cultural dimensions, and later expanded to six dimensions. 
Hofstede’s Cultural Dimensions Theory serves as a framework utilized to comprehend cultural disparities among countries. A key component of this theory is the Hofstede Value Survey,\footnote{\url{https://geerthofstede.com/research-and-vsm/vsm-2013/}} a questionnaire designed to comparing cultural values and beliefs of similar individuals between cultures.
These dimensions are \textit{Power Distance} (\textit{pdi}): reflect the acceptance of unequal power distribution in a society; \textit{Individualism} (\textit{idv}): examine the level of interdependence and self-definition within a society; \textit{Masculinity} (\textit{mas}): explore the emphasis on competition and achievement (Masculine) or caring for others and quality of life (Feminine); \textit{Uncertainty Avoidance} (\textit{uai}): deal with a society's response to ambiguity and efforts to minimize uncertainty, \textit{Long-Term Orientation} (\textit{lto}): describe how societies balance tradition with future readiness; and \textit{Indulgence} (\textit{ivr}): focuse on the control of desires and impulses based on cultural upbringing. 
These cultural dimensions provide insights into various aspects of a society's values and behaviors, help to understand the cultural variations and preferences within societies and how they shape attitudes, behaviors, and priorities.

We symbolize these features as $C=\left\{ c_i \right\}$, where $i\in \left\{ pdi, idv, mas, uai, lto, ivr \right\} $, with each feature value ranging from 0 to 100.\footnote{The cultural dimension values for each country can be found in \url{https://geerthofstede.com/research-and-vsm/dimension-data-matrix/}.}
To characterize the differences between the transfer dataset and the target dataset in different cultural dimensions, we represent them using the proportional values of each cultural dimension, which is defined as $Rc_i={{c_{i}^{\mathrm{tsf}}}/{c_{i}^{\mathrm{tgt}}}}$.
These cultural distances are denoted as $\textit{CulDim} = \{Rc_i\}$.

\subsection{Domain-Specific Feature}
% Apart from cases of anti-bias, the detection of offensive language is largely associated with the use of offensive terms in a statement. There are differences between cultures in terms of what is considered offensive vocabulary and expressions.

% The proliferation of offensive language on the internet has raised concerns about its negative impact on users' well-being and online safety. In response, researchers have been developing methods to detect and mitigate offensive language, especially in the context of transfer learning for low-resource languages. One promising approach is to analyze domain-specific features related to offensive language. In this study, we explore the potential of using offensive language distance as a feature to predict the success of transfer learning for Offensive Language Detection (OLD) tasks.

% To assess the effectiveness of transfer learning 

To predict transfer learning success in OLD tasks, we explore using offensive word distance as a feature, inspired by \citet{sun-etal-2021-cross} who introduce Emotion Semantics Distance (ESD) for measuring lexical similarities of emotions across languages. 
In contrast to their method of calculating distances using 24 emotion concepts~\cite{jackson2019emotion}, we calculate offensive distance based on the alignment of offensive concepts across languages.

Specifically, we utilize \href{https://github.com/valeriobasile/hurtlex/tree/master}{Hurtlex}, a multilingual lexicon comprising offensive, aggressive, and hateful words over 50 languages~\cite{bassignana2018hurtlex}.\footnote{License \url{https://creativecommons.org/licenses/by-nc-sa/4.0/}} This resource offers over 1000 aligned offensive lemmas across languages. For instance, words related to the concept of `clown' in various languages (i.e. klaun, klown, bouffon, payaso, jokar, Kōmāḷi, \begin{CJK}{UTF8}{gbsn}{小丑}\end{CJK}, etc.) are grouped together.
%as aligned. 
% We select 16 languages, and for every word in the aligned table-lexicon, we get the aligned FastText embeddings.
% %\footnote{\url{https://fasttext.cc/docs/en/aligned-vectors.html}}
% \cite{joulin-etal-2018-loss}.\footnote{The language selection is based on the OLD, DEP and SA NLP tasks following \citet{sun-etal-2021-cross}. For Japanese and Tamil we take the average distance for each language since there are no aligned Japanese embeddings with FastText, and Tamil is not in Hurtlex.} 
We obtain aligned word embeddings based on FastText~\cite{joulin-etal-2018-loss},\footnote{\url{https://fasttext.cc/docs/en/aligned-vectors.html}} and average the distance between all aligned offensive concepts between language pairs. 
Mathematically, the offensive distance $\textit{OffDist}_{(\text{tsf}, \text{tgt})}$ from transfer language to 
target language is defined as follows:

% calculate the pair-wise distance between the aligned word embeddings of each language using cosine similarity and average over the set of offensive word concepts. We use this metric similarly to \citet{sun-etal-2021-cross} as it considers the angle between vectors rather than their magnitude, making it more suitable for high-dimensional spaces like word embeddings. 
% We define \textit{OffDist} as follows:

% \[
% \textit{OffDist}\left( \text{tsf},\text{tgt} \right) =\left( \sum_{w\in \mathcal{W}}{\frac{\boldsymbol{v}_{w}^{\text{tsf}}\cdot \boldsymbol{v}_{w}^{\text{tgt}}}{\left\| \boldsymbol{v}_{w}^{p} \right\| \cdot \left\| \boldsymbol{v}_{w}^{q} \right\|}} \right) /\left| \mathcal{W} \right|
% \]
% \[
% \textit{OffDist}_{(\text{tsf}, \text{tgt})} =  \sum_{w \in W}\cos(\boldsymbol{v}_{(\text{tsf}, \text{w})}, \boldsymbol{v}_{(\text{tgt}, \text{w})}) \slash {|W|} 
% \]
\[
\textit{OffDist}_{(\text{tsf}, \text{tgt})}=\sum_{w\in \mathcal{W}}{\cos \left( \boldsymbol{v}_{w}^{\text{tsf}},\boldsymbol{v}_{w}^{\text{tgt}} \right)} /\left| \mathcal{W} \right|
\]
where $\boldsymbol{v}_{w}^{\text{tsf}}$ and $\boldsymbol{v}_{w}^{\text{tgt}}$ respectively represent the aligned word vectors of offensive aligned word $w$ in FastText for transfer language and target language, $\mathcal{W} =\left\{ \left( w^{\text{tsf}},w^{\text{tgt}} \right) \right\}$ is the set of all offensive aligned words between transfer and target languages. 
In our example, `clown' is the offensive word concept belonging to $\mathcal{W}$. 
Assuming transfer language is English and target language is Czech, then $w^{\text{tsf}}$ and $w^{\text{tgt}}$ would be the words `clown' and `klaun',  $\boldsymbol{v}_{w}^{\text{tgt}}$ and $\boldsymbol{v}_{w}^{\text{tgt}}$ denote the corresponding word embeddings extracted with Fasttext respectively.
% Assuming the transfer laguage is English and the target language is Czech, then \text{w} would be the words `clown' and `klaun' and \textbf{v} the word embeddings extracted with Fasttext respectively.

% \[
% \textit{OffDist}_{(\text{p}, \text{q})} =  \sum_{w \in W}\cos(\textbf{v}_{(\text{p}, \text{w})}, \textbf{v}_{(\text{q}, \text{w})}) \slash {|W|} 
% \]
% where $W$ is the set of offensive word concepts, and \textbf{v} represents the Fasttext aligned word vectors of the offensive word $w$ between the transfer $tfs$ and the target languages $tgt$. 

\begin{table*}[ht]
\centering
\scalebox{0.72}{
\begin{tabular}{llllllr}
\toprule
\textbf{Dataset}  & \textbf{Language} & \textbf{Country} & \textbf{Region} & \textbf{Offensive Label} & \textbf{Source} & \textbf{Size}    \\ \midrule
% ArabicOLD  \small{\cite{mubarak-etal-2021-arabic}}  & Arabic & -    &  Middle East  & Vulgar; Hate Speech     &  Tweet  & 10000 \\
COLD \small{\cite{deng-etal-2022-cold}}    & Chinese     & China    &  East Asia        & Offensive     & \multirow{2}{1.8cm}{Zhihu, Weibo}      & 37480 \\ \\
ChileOLD  & Spanish     & Chile      & South     & \multirow{3}{5cm}{intentional profanity/vulgarity; unintended profanity/vulgarity; insult/appellation; hate speech}           & Tweet      & 9834                          \\\small {\cite{arango-monnar-etal-2022-resources}}&&&America \\\\ 
DeTox \small{\cite{demus-etal-2022-comprehensive}} & German & Germany   & West Europe   & Hate Speech
% \multirow{7}{5cm}{Job; Political Attitude; Personal Engagement and Interests; Sexual Identity; Physical, Psychological or Mental Characteristics; Nationality; Religion; Social Status; World View; Ethnicity} 
& Tweet   & 10278 \\ 
HinHD  \small{\cite{mandl2019overview}}  & Hindi & India    &  South Asia  & \multirow{2}{5cm}{Offensive; Hate Speech; Defamation}     &  \multirow{2}{1.8cm}{Twitter, Facebook, WhatsApp}   & 8192 \\ \\ \\
KOLD \small{\cite{jeong-etal-2022-kold}}    & Korean     & South Korea    &  East Asia        & Offensive     & \multirow{2}{1.8cm}{NAVER, YouTube}       & 40429 \\ \\
NJH-UK \small{\cite{bianchi-etal-2022-just}}  & English       & United    & West Europe     & \multirow{3}{5cm}{Outrage; Insults; Profanity; Character Assassination; Discrimination; Hostility}                & Tweet        &   11190                   \\ &&Kingdom\\
NJH-US \small{\cite{bianchi-etal-2022-just}}  & English       & United States    & North America     &             & Tweet        &    9086                  \\ 
% OLID \small{\cite{zampieri-etal-2019-predicting, zampieri-etal-2019-semeval}}   &  English      & -    &  -        & Offensive    & Tweet      & 14100\\ 
PolEval \small{\cite{publ152265}}   &  Polish      & Poland    &  East Europe        & Cyberbullying; Hate Speech    & Tweet      & 10041\\ 
TurkishOLD  \small{\cite{coltekin-2020-corpus}}   &  Turkish    & Turkey    & Middle East         & Offensive    & Tweet & 34792 \\
% SOLD \small{\cite{ranasinghe2022sold}}     &  Sinhala      &  Sri Lanka   &  South Asian        & Offensive     &  Tweet & 10000 \\
% DKHATE \small{\cite{sigurbergsson-derczynski-2020-offensive}}   & Danish    &  Denmark   & North Europe         & Offensive     & \multirow{3}{1.8cm}{Tweet, Facebook, Reddit}     & 3290 \\ \\ \\
    % &  Bulgarian    &  Bulgaria    &          &      &       & \\
% CAD \small{\cite{vidgen-etal-2021-introducing}} & English  &     &          &      &       & \\
    % &        &     &          &      &       & \\
\bottomrule
\end{tabular}}
\caption{Statistics of OLD datasets used in our experiments, covering 8 languages and 9 countries.}
\label{tab:datasets}
\end{table*}

\subsection{Features Analysis}
% We conduct a visual evaluation of the proposed features that are relevant to the culture. 
To analyze and qualitatively evaluate our proposed features we attempt to visualize them.
% \textit{CulDim} features are country-level, while \textit{OffDist} is language-level. 
For \textit{OffDist} (language-level), we adopt the language selection from \citet{sun-etal-2021-cross}, including 16 languages. As for \textit{CulDim} (country-level), we select 16 countries based on their relevance to the experimental OLD datasets and the chosen languages.\footnote{The list of selected countries and languages, along with their corresponding abbreviations in Figure~\ref{fig:analysis_cultural} and Figure~\ref{fig:analysis_off}, can be found in Appendix~\ref{app:CL}.}

\paragraph{Cultural dimensions.}
These dimensions provide insights into societal perceptions and interactions across various aspects of life. To visualize the relationships between countries' cultural values, we use T-SNE in Figure~\ref{fig:analysis_cultural}, where the proximity of points on the plot indicates the degree of similarity in their cultural values.
Observations from Figure~\ref{fig:analysis_cultural} reveal that there is a degree of consistency between cultural similarity and geographical proximity. 
For instance, countries in Western Europe and the Middle East tend to be closer in Figure~\ref{fig:analysis_cultural}.
However, geography alone doesn't solely determine cultural similarity. Notably, the United States and the United Kingdom are visually close despite their geographic distance, while Japan and Germany, though geographically distant, show cultural proximity. This illustrates the intricate interplay between culture and geography, transcending physical boundaries.

% These dimensions provide insights into how societies and individuals within them perceive and interact with various aspects of life.
% To explore the relationships between countries based on their cultural values, we employ T-SNE for visualization in Figure~\ref{fig:analysis_cultural}. 
% The proximity of points on the plot indicates the degree of similarity in their cultural values. 
% Countries that are closer together share more similar cultural dimensions, while those that are farther apart exhibit greater dissimilarities.
% Observations from Figure~\ref{fig:analysis_cultural} reveal that there is a degree of consistency between cultural similarity and geographical proximity. 
% For instance, countries in Western Europe and the Middle East tend to be closer in the Figure~\ref{fig:analysis_cultural}. However, it is important to note that geographic location alone does not serve as an absolute criterion for measuring cultural similarity. An interesting finding is that despite being geographically distant, the United States and the United Kingdom exhibit close proximity in the visualization. Conversely, Japan and Germany, although not geographically close, demonstrate cultural proximity. This suggests that cultural influences can transcend geographic boundaries, highlighting the complex interplay between culture and geography.

\paragraph{Offensive distance.}
Following \citet{sun-etal-2021-cross}, we create a language network based on offensive distances, shown as Figure~\ref{fig:analysis_off}.\footnote{Japanese and Tamil are excluded due to lack of aligned FastText embeddings and a Hurtlex lexicon, respectively.} Each node corresponds to a language, distinguished by its cultural area. Languages are sorted based on offensive distance, and an edge is drawn between languages that rank among the top-2 closest.
Based on this network we can see that the offensive word distances seem to follow a language proximity pattern. Some clusters seem to appear for the Romance and Slavic languages. 
75.8\% of the edges correspond to cultural areas, matching exactly the ESD network and surpassing the syntactic network~\cite{sun-etal-2021-cross}. Emotion and offensive word similarity therefore seem to have a similar relationship to cultural areas.

% This feature refers to the cosine distance between cross-lingual aligned embeddings of offensive words. We visualize the cross-lingual distances as a network following \citet{sun-etal-2021-cross}. 

% There also seem to be a lot of languages connected to English, which can be explained by the fact that Hurtlex multilingual lexica have a lot of English offensive words(e.g. the Chinese lexicon includes words written in English such as twerp, nerd, gay, blooper). 

% Lastly, some edges between languages make little sense, e.g., the edge between Hindi and Chinese or German and Korean. A possible explanation could be that using embeddings alone for alignment may not capture fine-grained relationships between languages \cite{alaux2018unsupervised}.

\paragraph{Culture vs. geography.}
Based on the above analysis, whether visualizing countries based on \textit{CulDim} or visualizing languages based on \textit{OffDist}, they can approximately fit the grouping based on geographical location but do not strictly match it. This highlights that geographical differences cannot be absolute indicators of cultural distinctions. 
% Cultural differences can exist within a geographical region, and people living in the same geographic area can have diverse cultural identities.
Culture is a complex and multifaceted concept that is influenced by various factors beyond geography, such as history, politics, migration, and globalization~\cite{taras2016does}.
Thus, exploring additional culturally relevant features becomes necessary.
Appendix~\ref{app:correlation} presents the feature correlation analysis involved in the paper.

% the relation between cultural difference and geography

% appendix show more analysis about cultural dimension and different features

%--------------------------------------
\begin{table*}[ht]
\centering
\scalebox{0.85}{
\begin{tabular}{@{}lllllrrllrr@{}}
\cmidrule(r){1-3} \cmidrule(lr){5-7} \cmidrule(l){9-11}
% \toprule
                           & \multicolumn{1}{r}{\textbf{MAP}} & \multicolumn{1}{r}{\textbf{NDGC}} & \multirow{9}{*}{} &                            & \textbf{MAP}   & \textbf{NDGC}  & \multirow{9}{*}{} &                            & \textbf{MAP}   & \textbf{NDGC}  \\ \cmidrule(r){1-3} \cmidrule(lr){5-7} \cmidrule(l){9-11} 
\textbf{LangRank}          & 47.48                            & 59.33                              &                   & \textbf{MTVEC}             & 51.67          & 68.23          &                   & \textbf{Colex2Lang}        & 57.78          & 71.41          \\ \cmidrule(r){1-3} \cmidrule(lr){5-7} \cmidrule(l){9-11} 
\quad \textit{\small \textbf{+PRAG}}    & 50.65                            & 53.60                              &                   & \quad \textit{\small \textbf{+PRAG}}    & 56.98          & 73.87          &                   & \quad \textit{\small \textbf{+PRAG}}    & 55.77          & 71.95          \\
\quad \textit{\small \textbf{+OffDist}} & 51.08                            & 60.17                              &                   & \quad \textit{\small \textbf{+OffDist}} & 64.26          & 76.50          &                   & \quad \textit{\small \textbf{+OffDist}} & 59.64          & 73.81          \\
\quad \textit{\small \textbf{+CulDim}}  & 51.23                            & 60.71                              &                   & \quad \textit{\small \textbf{+CulDim}}  & 70.28          & 76.92          &                   & \quad \textit{\small \textbf{+CulDim}}  & {\underline{73.25}}    & 71.56          \\ \cmidrule(r){1-3} \cmidrule(lr){5-7} \cmidrule(l){9-11} 
\quad \textit{\small \textbf{+P+O}}     & 50.90                            & 58.35                              &                   & \quad \textit{\small \textbf{+P+O}}     & 49.46          & 62.30          &                   & \quad \textit{\small \textbf{+P+O}}     & 55.75          & 73.26          \\
\quad \textit{\small \textbf{+P+C}}     & \textbf{57.45}                   & \textbf{67.63}                     &                   & \quad \textit{\small \textbf{+P+C}}     & {\underline{74.44}}    & {\underline{76.53}}    &                   & \quad \textit{\small \textbf{+P+C}}     & 69.63          & {\underline{76.33}}    \\
\quad \textit{\small \textbf{+C+O}}     & 53.45                            & 61.47                              &                   & \quad \textit{\small \textbf{+C+O}}     & 73.33          & 73.87          &                   & \quad \textit{\small \textbf{+C+O}}     & \textbf{76.69} & \textbf{79.34} \\ \cmidrule(r){1-3} \cmidrule(lr){5-7} \cmidrule(l){9-11} 
\quad \textit{\small \textbf{+P+C+O}}   & {\underline{56.54}}                      & {\underline{62.73}}                        &                   & \quad \textit{\small \textbf{+P+C+O}}   & \textbf{76.67} & \textbf{79.61} &                   & \quad \textit{\small \textbf{+P+C+O}}   & 68.99          & 72.06          \\ 
% \bottomrule
\cmidrule(r){1-3} \cmidrule(lr){5-7} \cmidrule(l){9-11} 
\end{tabular}}
\caption{The performance of transfer language prediction. \textbf{Bold} for best performance, \underline{underline} for second best performance. \textit{P}, \textit{O} and \textit{C} stand for \textit{PRAG}, \textit{OffDist}, and \textit{CulDim} respectively.
}
\label{tab:overall_result}
\end{table*}

\begin{table}[ht]
\centering
\scalebox{0.75}{
\begin{tabular}{l|lll}
\toprule
\textbf{Target Datasets} & \multicolumn{3}{c}{\textbf{Ranking comparison}} \\
\midrule
\multirow{2}{*}{\textbf{COLD}}          & \increasebg{\small \blue{KOLD}}       & \increasebg{\small \blue{NJH\_US}}        & \increasebg{\small NJH\_UK}    \\
                                        & \small TurkishOLD        & \small \blue{KOLD}           & \small \blue{NJH\_US} \\\midrule
\multirow{2}{*}{\textbf{ChileOLD}}      & \increasebg{\small \blue{NJH\_UK}}    & \increasebg{\small Hindi}                 & \increasebg{\small NJH\_US}    \\
                                        & \small KOLD              & \small \blue{NJH\_UK}        & \small COLD \\  \midrule
\multirow{2}{*}{\textbf{DeTox}}         & \increasebg{\small \blue{TurkishOLD}} & \increasebg{\small \blue{KOLD}}           & \increasebg{\small \red{ChileOLD}}     \\
                                        & \small \blue{KOLD}       & \small \blue{TurkishOLD}     & \small \red{ChileOLD}     \\  \midrule
\multirow{2}{*}{\textbf{Hindi}}         & \increasebg{\small \blue{ChileOLD}}   & \increasebg{\small \blue{NJH\_UK}}        & \increasebg{\small \blue{NJH\_US}}      \\
                                        & \small \blue{NJH\_UK}    & \small \blue{NJH\_US}        & \small \blue{ChileOLD}    \\  \midrule
\multirow{2}{*}{\textbf{KOLD}}          & \increasebg{\small \blue{NJH\_UK}}    & \increasebg{\small COLD}                  & \increasebg{\small Hindi}         \\
                                        & \small NJH\_US           & \small \blue{NJH\_UK}        & \small ChileOLD      \\    \midrule
\multirow{2}{*}{\textbf{NJH\_UK}}       & \increasebg{\small \blue{NJH\_US}}    & \increasebg{\small \blue{COLD}}           & \increasebg{\small \red{Hindi}}        \\
                                        & \small \blue{COLD}       & \small \blue{NJH\_US}        & \small \red{Hindi}     \\  \midrule
\multirow{2}{*}{\textbf{NJH\_US}}       & \increasebg{\small {NJH\_UK}}         & \increasebg{\small \blue{COLD }  }        & \increasebg{\small \blue{Hindi } }     \\
                                        & \small \blue{COLD}       & \small \blue{Hindi}          & \small KOLD      \\     \midrule
\multirow{2}{*}{\textbf{PolEval}}       & \increasebg{\small \blue{ChileOLD}}   & \increasebg{\small TurkishOLD}            & \increasebg{\small \blue{KOLD}}    \\
                                        & \small \blue{KOLD}       & \small \blue{ChileOLD}       & \small NJH\_UK   \\    \midrule
\multirow{2}{*}{\textbf{TurkishOLD}}    & \increasebg{\small \red{ChileOLD}}    & \increasebg{\small \red{NJH\_US}}         & \increasebg{\small \red{KOLD}}   \\
                                        & \small \red{ChileOLD}    & \small \red{NJH\_US}         & \small \red{KOLD} \\
\bottomrule
\end{tabular}}
\caption{Comparison of Top-3 Transfer Datasets. The first line represents \lightred{ground truth rankings}, while the second line represents predicted rankings. Transfer datasets with exact rankings are highlighted in \red{red}, and those predicted within the top-3 positions are highlighted in \blue{blue}. }
\label{tab:error_analysis}
\end{table}

% \begin{table}[]
% \begin{tabular}{llll}
% \hline
%                    & \multicolumn{3}{l}{haha} \\ \hline
% \multirow{2}{*}{a} & 1      & 1      & 1      \\
%                    & 2      & 2      & 2      \\ \hline
% \multirow{2}{*}{b} & 3      & 3      & 3      \\
%                    & 4      & 4      & 4      \\ \hline
% \end{tabular}
% \caption{}
% \label{tab:my-table}
% \end{table}

\section{Can we predict transfer languages for OLD successfully?}
\subsection{Experimental Settings}

\paragraph{OLD datasets.} 
To acquire country-level cultural dimension features, we carefully select 9 OLD datasets that include annotators' nationality information, ensuring the availability of country context.
The OLD dataset statistics are shown in Table~\ref{tab:datasets} (refer to Appendix~\ref{app:Datasets} for more details). 
Offensive language encompasses various aspects~\cite{bianchi-etal-2022-just, demus-etal-2022-comprehensive}, including hate speech, insults, threats, and discrimination. 
In this paper, we simplify the detection tasks in the OLD datasets into binary classification tasks, considering a text offensive if it is labeled with any offensive label.

\begin{figure}[ht]
\centering
\includegraphics[width=1\linewidth]{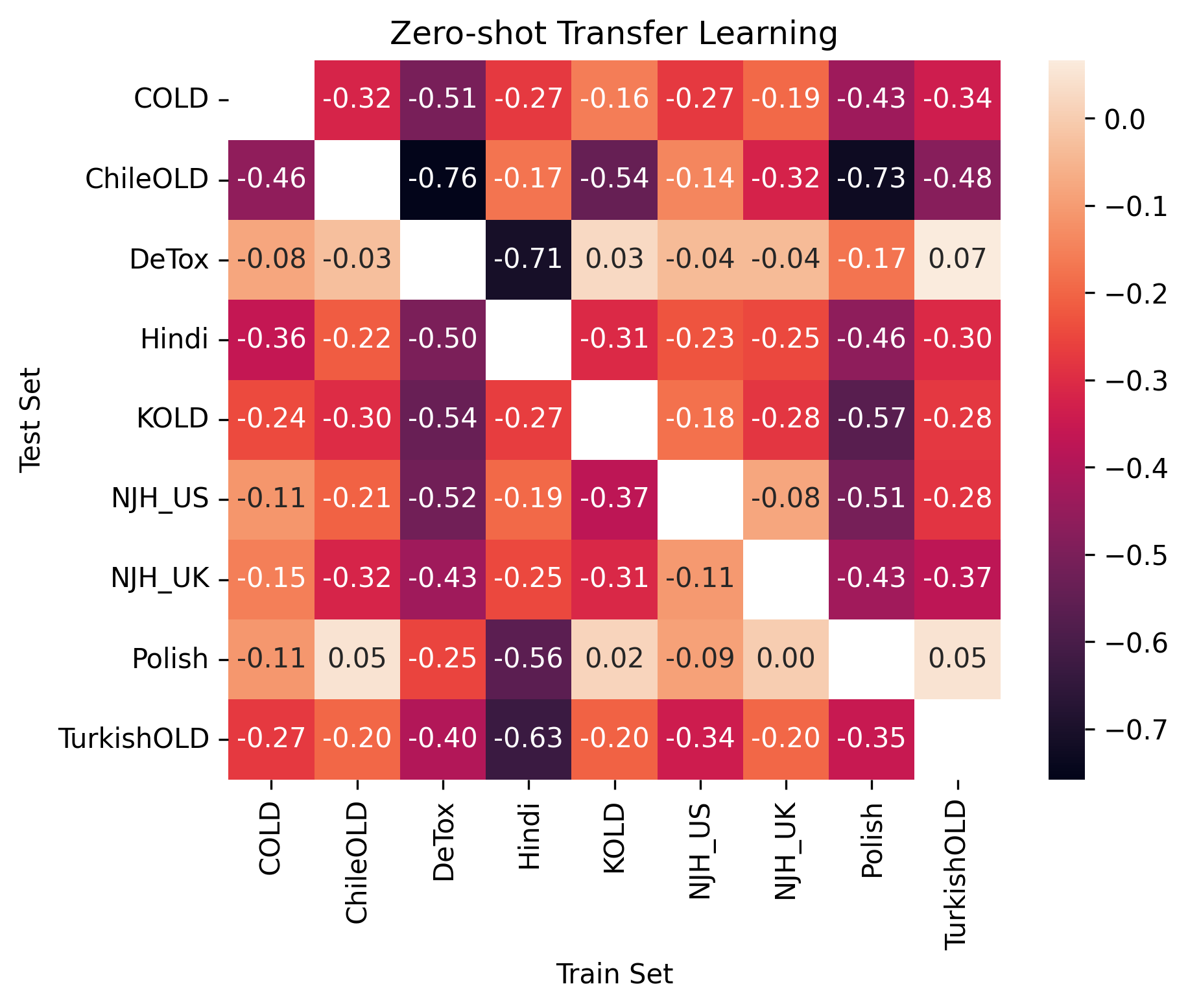}
\caption{Relative losses (in percent) from zero-shot transfer learning models over the intra-cultural models.}
\label{fig:TL_LLM}
\end{figure}

\paragraph{Zero-shot transfer.} 
We train intra-cultural XLM-R~\cite{conneau-etal-2020-unsupervised} for each dataset to ensure cultural consistency. Then, we perform zero-shot transfer (ZST) using XLM-R~\footnote{\url{https://huggingface.co/xlm-roberta-base}} on 72 directed dataset pairs, simulating cross-cultural implementation. 
% We fine-tune XLM-R on the transfer dataset $\mathrm{D}^{\mathrm{tsf}}$ and measure performance on the testset of the target dataset $\mathrm{D}^{\mathrm{tgt}}$ using macro F1 score. 
Specifically, for all 72 directed dataset pairs, we fine-tune the XLM-R model on the transfer dataset with learning rate 1e-5. The training epoch is 20, training batch size is 8, we adopt the early-stopping method with patience 10 on the target dataset’s dev set. In order to mitigate the influence of randomness in our experiments and ensure result reliability, we run five times for each dataset pair. Subsequently, we compute the average performance (macro F1 score) of the trained model on the target dataset's test set, representing the transfer performance for each dataset pair.
Figure~\ref{fig:TL_LLM} illustrates the relative losses incurred by ZST compared to intra-cultural models, highlighting the varying impacts of different transfer datasets on the same target dataset. This presents a challenge in selecting appropriate transfer datasets for low-resource target datasets without prior implementation. These findings underscore the importance of predicting transfer learning success. Lastly, we obtain the ground truth ranking results for the transfer of OLD based on the experimental results of ZST.

\paragraph{Transfer ranking prediction.}
Following \citet{sun-etal-2021-cross}, we use LightGBM~\cite{ke2017lightgbm} as the ranking model, known for its effectiveness in ranking tasks. We evaluate the performance using Mean Average Precision (MAP) and Normalized Discounted Cumulative Gain (NDCG) metrics. MAP calculates the average precision across all ranks, while NDCG considers both position and relevance. For our analysis, we focus on the top 3 ranked items for both metrics.

\paragraph{Baseline features.}
We employ the following ranking baselines: LangRank~\cite{lin-etal-2019-choosing}, incorporating 13 features to rank items, including data-dependent and linguistic-dependent features; MTV{\small EC}~\cite{malaviya-etal-2017-learning}, learning 512-dimensional language vectors to infer syntactic, phonological, and phonetic inventory features; Colex2Lang~\cite{chen-etal-2023-colex2lang}, learning language representations with node embedding algorithms based on a constructed large-scale synset. Additionally, we incorporate \textit{PRAG} features~\cite{sun-etal-2021-cross}, capturing cross-cultural linguistic patterns and assessing language pragmatics. Appendix~\ref{app:feature_group} provides a summary of the features used in each baseline and subsequent feature groups.

\subsection{Overall Results}
We examine the effectiveness of proposed features in predicting transfer dataset rankings for OLD~\footnote{The specific cultural dimension features used in our experiments are shown in Appendix~\ref{app:Cultural_value}.}.
We specifically incorporate different features and their combinations into the baseline features, and the experimental results are presented in Table~\ref{tab:overall_result}.
Results show that our proposed features,
\textit{CulDim} and \textit{OffDist}, independently enhance baseline performance, indicating the influence of cultural-value differences and domain-specific distance on transfer prediction. 
Moreover, Combining \textit{CulDim} consistently improves ranking prediction, reinforcing the reliability of cross-cultural features in culture-loaded task transfer. 
Surprisingly, the MTV{\small EV} group surpasses the LangRank group, despite lacking data-dependent features. This underscores the impact of culture-dependent and language-dependent features on accurate and robust transfer learning predictions for culture-loaded tasks.

Moreover, using the best results from Table~\ref{tab:overall_result}, we visualize the top-3 transfer datasets for each target dataset in Table~\ref{tab:error_analysis}. By comparing the ground truth rankings with the predicted rankings, we further assess prediction accuracy. 
The analysis reveals that for TurkishOLD, the top-3 transfer datasets, including their specific rankings, are accurately predicted.  
When considering only the prediction accuracy for the top-3 rankings without considering the specific positions, the accuracy rate is 100\% for four datasets and 67\% for three datasets.
Except for NJH\_US, the highest-ranked transfer datasets are accurately predicted to be within the top-3 for other datasets. These results demonstrate the potential of our proposed features in reducing trial and error in migration data selection. Accurate prediction of top-3 transfer datasets empowers researchers and practitioners to make informed decisions, enhancing the efficiency and effectiveness of cross-cultural transfer learning.

\begin{table}[]
\centering
\scalebox{0.82}{
\begin{tabular}{@{}l|rrrrrr@{}}
\toprule
              & \multicolumn{2}{c}{\textbf{LangRank}} & \multicolumn{2}{c}{\textbf{MTVEC}} & \multicolumn{2}{c}{\textbf{Colex2Lang}} \\
              & \small \textbf{MAP}      & \small \textbf{NDGC}     &\small \textbf{MAP}    &\small \textbf{NDGC}    &\small \textbf{MAP}  & \small \textbf{NDGC}      \\ \midrule
\textbf{BEST} & 57.45             & 67.63             & 76.67           & 79.61            & 76.69              & 79.34              \\ \midrule
\quad\textbf{-pdi} & 51.52             & 62.46             & 76.28           & 76.31            & 76.34              & 78.95              \\
\quad\textbf{-idv} & 51.98             & 65.18             & 74.81           & 78.52            & 72.06              & 75.86              \\
\quad\textbf{-mas} & 54.09             & 63.20             & 70.94           & 72.00            & 68.92              & 72.74              \\
\quad\textbf{-uai} & 46.12             & 59.55             & 76.76           & 77.89            & 76.06              & 80.96              \\
\quad\textbf{-lto} & 51.19             & 67.07             & 61.94           & 67.16            & 71.80              & 74.36              \\
\quad\textbf{-ivr} & 55.30             & 61.95             & 73.33           & 75.61            & 69.95              & 74.13              \\ \bottomrule
\end{tabular}}
\caption{Ablation analysis for the six cultural dimensions, demonstrating that all dimensions contribute to the ranking performance and are therefore meaningful for offensive language detection generally.}
\label{tab:ablation_study}
\end{table}

\begin{table*}[ht]
\centering
\scalebox{0.66}{
\begin{tabular}{@{}l|rr|rr|rr|ll|rr|rr|rr|rr@{}}
\toprule
\multirow{2}{*}{\textbf{Datasets}} & \multicolumn{2}{c|}{\textbf{Data-specific}} & \multicolumn{2}{c|}{\textbf{Typology}} & \multicolumn{2}{c|}{\textbf{Geography}} & \multicolumn{2}{l|}{\textbf{Orthography}}                             & \multicolumn{2}{c|}{\textbf{Pragmatic}} & \multicolumn{2}{c|}{\textbf{PRAG}} & \multicolumn{2}{c|}{\textbf{OFF}} & \multicolumn{2}{c}{\textbf{Cultural}} \\
                                   & \small \textbf{MAP}         & \small \textbf{NDGC}        & \small \textbf{MAP}      & \small \textbf{NDGC}      & \small \textbf{MAP}       & \small \textbf{NDGC}      & \multicolumn{1}{r}{\small \textbf{MAP}} & \multicolumn{1}{r|}{\small \textbf{NDGC}} & \small \textbf{MAP}       & \small \textbf{NDGC}      & \small \textbf{MAP}    & \small \textbf{NDGC}    & \small \textbf{MAP}    & \small \textbf{NDGC}   & \small \textbf{MAP}      & \small \textbf{NDGC}     \\ \midrule
\textbf{COLD}                      & 75.00                & 69.58                & 25.00             & 15.79              & 26.67              & 38.21              & 58.33                            & 82.07                              & 33.33              & 65.10              & 45.00           & 79.24            & 24.29           & 9.55            & 58.33             & 73.11             \\
\textbf{ChileOLD}                  & 39.29                & 60.83                & 20.83             & -3.31              & 29.17              & 57.31              & 29.17                            & 28.07                              & 26.79              & 31.59              & 26.79           & 31.38            & 26.79           & 15.10           & 45.00             & 78.76             \\
\textbf{DeTox}                     & 22.62                & 31.47                & 26.79             & 47.53              & 24.29              & 35.38              & 33.33                            & 59.26                              & 26.67              & 27.37              & 26.67           & 22.11            & 66.67           & 77.78           & 62.5              & 76.35             \\
\textbf{Hindi}                     & 83.33                & 94.34                & 25.00             & 23.59              & 25.00              & 38.21              & 41.67                            & 77.00                              & 62.50              & 68.41              & 41.67           & 56.35            & 20.83           & 35.02           & 30.95             & 42.69             \\
\textbf{KOLD}                      & 64.29                & 79.72                & 32.5              & 21.24              & 26.67              & 38.21              & 37.50                            & 67.45                              & 41.67              & 42.69              & 50.00           & 55.18            & 66.67           & 79.72           & 30.95             & 23.59             \\
\textbf{NJH\_UK}                   & 41.67                & 48.35                & 64.29             & 69.58              & 29.17              & 76.41              & 45.00                            & 70.28                              & 83.33              & 94.34              & 100.00          & 96.69            & 62.50           & 80.31           & 66.67             & 79.72             \\
\textbf{NJH\_US}                   & 36.67                & 79.72                & 62.5              & 71.93              & 29.17              & 63.68              & 33.33                            & 54.59                              & 75.00              & 71.93              & 75.00           & 71.93            & 66.67           & 70.17           & 39.29             & 59.66             \\
\textbf{PolEval}                   & 32.06                & 42.38                & 33.13             & 29.12              & 45.24              & 51.13              & \multicolumn{1}{r}{63.89}        & \multicolumn{1}{r|}{55.82}         & 68.06              & 71.51              & 56.94           & 63.98            & 45.24           & 54.26           & 100.00            & 99.01             \\
\textbf{TurkishOLD}                & 26.79                & 52.83                & 70.00             & 67.24              & 40.00              & 53.49              & \multicolumn{1}{r}{24.29}        & \multicolumn{1}{r|}{24.17}         & 66.67              & 79.72              & 32.50           & 65.21            & 22.50           & 55.66           & 50.00             & 46.21             \\ \midrule
\textbf{AVG.}                      & 46.86                & 62.14                & 40.00             & 38.08              & 30.6               & 50.23              & \multicolumn{1}{r}{40.72}        & \multicolumn{1}{r|}{57.63}         & \textbf{53.78}              & \underline{61.41}              & 50.51           & 60.23            & 44.68           & 53.06           & \underline{53.74}             & \textbf{64.34}             \\ \bottomrule
\end{tabular}}
\caption{Performance comparison of OLD ranking predictions. \textbf{Bold} for best performance, \underline{underline} for second best performance.}
\label{tab:new_baseline}
\end{table*}

\subsection{Cultural Dimensions Analysis}
To examine the influence of each cultural dimension on offensive language detection, we perform an ablation experiment using the optimal feature combination from each baseline setting in Table~\ref{tab:overall_result}, all of which include \textit{CulDim}. The results in Table~\ref{tab:ablation_study} demonstrate that almost each cultural dimension can contribute individually to the prediction of OLD transfer. These contributions exhibit variation depending on the specific baseline features utilized. 
It's important to note that cultural dimensions are not mutually exclusive, and a given culture can exhibit varying degrees of each dimension. 
Furthermore, the relationship between culture and OLD is complex and can be influenced by factors beyond these dimensions, such as historical context, social norms, and individual differences.

However, the cultural dimension values are somewhat correlated with the sensitivity of offensive language detection.
~\citet{khan2014impact} states that ``High power distance employees will be less likely to perceive their supervisory abusive'' and ``High individualistic cultural employees will be more likely to perceive their supervisory abusive''.
~\citet{bond1985responses} find that cultural variations (including Cultural Collectivism and Power Distance) in the use of social control are related to perceptions of an insult and of the insulter.
~\citet{van2008terms} also highlight that people from different cultures exploit different categories of verbal abuse.
Our experimental results further substantiate these research findings.

% For example, in an individualistic culture with a high score in the \texttt{Individualism} dimension, OLD systems may focus more on identifying direct personal attacks or threats rather than policing more general offensive language, while the detection of language that disrupts social harmony, denigrates specific groups, or violates cultural norms and values takes precedence in a collectivistic culture.
% The same cross-cultural differences in OLD are also reflected in the \texttt{Power distance} dimension. High power distance cultures may focus on the identification of disrespectful language towards authority figures, whereas such language is desalinated in low power distance cultures.

% ablation study, analysis
% Table~\ref{tab:ablation_study}

% \usepackage{booktabs}
% \usepackage[normalem]{ulem}

\section{What types of features are most important for different types of tasks?}

\subsection{Feature Fine-grained Comparison}
To gain a deeper insight into the key features contributing to the success of the optimal transfer dataset prediction in OLD task, we categorize the involved features into eight distinct groups: Data-specific, Topology, Geography, Orthography, Pragmatic, PRAG, OFF, and Cultural.
The first five groups align with the grouping proposed by \citet{sun-etal-2021-cross} in their previous work. 
In particular, the Pragmatic group consists of PRAG features as well as three data-dependent features. So we establish a separate group specifically dedicated to the PRAG features to ensure their distinct recognition and analysis.
We introduce two new feature groups: OFF, which incorporates our newly proposed feature \textit{OffDist}, specifically designed for OLD tasks, and Cultural, which encompasses the \textit{CulDim} feature capturing cultural dimensions.

The performance of ranking models trained with different feature groups for OLD is presented in Table~\ref{tab:new_baseline}. 
The results highlight the varying performance of different feature types. The Cultural group stands out as the best-performing group in both metrics, closely followed by the Pragmatic group.
This finding confirms that the proposed \textit{CulDim} feature, which captures cultural-value differences, holds high predictive power for OLD tasks.
While the feature \textit{OffDist} can provide some auxiliary predictive power when combined with other features, it does not possess independent prediction ability on its own, which points to the limitations of the domain-specific feature.
% This observation confirms that proposed \textit{CulDim} capturing cultural-value differences is highly predictive for OLD. 

\begin{table}[ht]
\centering
\setlength{\tabcolsep}{2pt}
\scalebox{0.88}{
\begin{tabular}{l|rrrrrr}
\toprule
                       & \multicolumn{2}{c}{\textbf{DEP}} & \multicolumn{2}{c}{\textbf{SA}} & \multicolumn{2}{c}{\textbf{OLD}}                                     \\
                       & \small\textbf{MAP}   & \small\textbf{NDCG}   & \small\textbf{MAP}   & \small\textbf{NDCG}  & \multicolumn{1}{l}{\small\textbf{MAP}} & \multicolumn{1}{l}{\small\textbf{NDCG}} \\
\midrule
\textbf{Data-specific} & 36.94          & 52.83           & \underline{68.00}          & \underline{85.40}          & 46.86                            & 62.14                             \\
\textbf{Typology}      & \textbf{58.12}          & \textbf{79.41}           & 49.90          & 60.70          & 40.00                            & 38.08                             \\
\textbf{Geography}     & 34.12          & \underline{67.93}           & 24.90          & 55.00          & 30.60                            & 50.23                             \\
\textbf{Orthography}   & 35.54          & 65.50           & 34.20          & 56.60          & 40.72                            & 57.63                             \\
\textbf{Pragmatic}     & \underline{54.37}          & 60.50           & \textbf{73.20}          & \textbf{88.00}          & \textbf{53.78}                            & \underline{61.41}                             \\
\textbf{PRAG}          & 44.45          & 61.11           & 41.81          & 59.58          & 50.51                            & 60.23                            \\
% \textbf{TTR}           & 44.08          & 67.77           & 67.91          & 83.81          & 53.04                            & 60.23                                \\
% \textbf{OFF}           & 44.20          & 70.19           & 31.72          & 50.04          & 44.68                            & 53.06                            \\
% \textbf{EMO}           & 38.40          & 55.81           & 27.74          & 47.40          & 47.50                            & 56.52                                 \\
\midrule
\textbf{Cultural}      & 34.47          & 49.23           & 55.80          & 79.75          & \underline{53.74}                            & \textbf{64.34}      \\        
\bottomrule
\end{tabular}}
\caption{Ranking performance comparison on different NLP tasks: dependency parsing (DEP), sentiment analysis (SA) and offensive language detection (OLD). \textbf{Bold} for best performance, \underline{underline} for second best performance.}
\label{tab:TASK}
\end{table}

\subsection{Comparison in Different Tasks}
To quantify the generalization of different feature groups across various NLP tasks, we compare their prediction performance among OLD, Sentiment Analysis (SA) and Dependency Parsing (DEP).
Among these tasks, OLD and SA are relatively subjective in nature, while DEP is highly objective, with labels directly derived from linguistic rules. 
Given that the features of the Cultural group rely on the country information of the dataset, and the SA and DEP tasks involve datasets that only provide language information~\cite{sun-etal-2021-cross}, we filter out the countries where the relevant language is the sole official language.\footnote{\href{https://en.wikipedia.org/wiki/List_of_official_languages_by_country_and_territory}{\texttt{List of official languages by country and territory}}} We then calculate the average cultural dimension values of these countries to obtain the cultural dimension value for the language itself.

The comparison results are presented in Table~\ref{tab:TASK}. The Topology group demonstrates the highest predictive power for the DEP task. This can be attributed to the inclusion of syntactic distance measurements between language pairs, which aligns with the inherent characteristics of the DEP task.
Although the Cultural group does not rank as the most predictive for the SA task, it still achieves significant results. When comparing the performance of the Cultural group across different tasks, it is observed that the performance for OLD and SA tasks is higher than that for the DEP task. This finding highlights the valuable contribution of cultural features to the success of transfer learning predictions for culture-loaded tasks.

\section{Conclusion}
We explored the feasibility of predicting transfer learning success, identifying informative features for transfer learning effectiveness, and examined the generalizability of these features across different NLP tasks. We have gained valuable insights into the potential of incorporating cultural sensitivity in language technologies: dimensions from cultural values surveys consistently improve transfer success prediction, and offensive language distance features provide a further improvement.

We recommend annotation projects to consider not only the balance between genres and age groups but also encompass cultures from different regions. Including documentation with information about the cultural background of the annotators enables better interpretation of annotations considering biases and cultural nuances, and promotes a comprehensive and inclusive approach to research and analysis fostering the transparency, usability, and suitability of the dataset.
This study highlights the need for culturally informed datasets and the importance of including cultural factors in cross-lingual transfer learning for OLD.

Future work will augment the cross-cultural transfer learning strategy by incorporating cultural features directly into models. Further analysis will also conduct a socially situated multi-cultural human evaluation on language model predictions to address subjective nuances that cannot be identified in reference-based evaluation on isolated sentences. Finally, we will further explore the correlation between features like offensive word distance, geography, and culture, to gain insights into their relationships and effectiveness in capturing offensive language patterns.

% To better explore the sensitivity of cultural differences in the OLD task, we suggest that future annotation work should give the annotator's description.

% \textcolor{blue}{In future work, it would be interesting to investigate the effectiveness of using value questionnaires as features and compare the performance of human responses versus language model responses. 
% Another future direction could include, understanding the correlation between distances measured using different features such as offensive word distance, geography, and value questionnaires and shedding light on their relationships and their effectiveness in capturing offensive language patterns.
% Lastly, we would like to explore the differences between one-to-one transfer and many-to-one transfer approaches can provide insights into the transferability of knowledge in offensive language detection. 
% }

% Chinese vs Ethnic Chinese)

% \textbf{RQ4}: If using features from value questionnaires, is it effective use human responses or LM responses?

% \textbf{RQ5}: Is there a difference between one-to-one transfer and many-to-one transfer?

% \textbf{RQ6}: Is it beneficial to use the cultural features as inputs to the OLD model, and how (as features or as auxiliary objectives)?

% \textbf{RQ7}: What is the correlation between distances measured with different features (offensive word distance, geography, value questionnaires)?

\section*{Limitations}

% The direct use of language as a cultural grouping is not rigorous. People who speak the same language may have different cultural backgrounds. (US vs UK, 
While this study provides valuable insights into several aspects of cross-cultural and cross-lingual transfer learning for offensive language detection, there are certain limitations to consider.

The cultural features we introduce certainly do not fully capture the diverse nuances and complexities of different cultural contexts. While more fine-grained than linguistic categories, countries are clearly not homogeneous cultural entities, and assuming they are such can be limiting and harmful to diversity and inclusion.
Furthermore, our study is based on a specific set of datasets, tasks, and language pairs, which may limit the generalizability of our findings to other languages and domains. 
We employ early stopping on the target development set, a practice commonly used in many machine learning settings, while ~\cite{kann-etal-2019-towards} state that hand-labeled development sets in the target language are often assumed to be unavailable in the case of zero-shot learning. Moreover, the two most related works~\cite{lin-etal-2019-choosing, sun-etal-2021-cross} do not mention the details about development set.

% The evaluation of transfer learning success is inherently subjective and dependent on the available evaluation metrics, which may not capture all aspects of offensive language detection performance.

Regarding the offensive distance feature, it is clear that using embeddings alone for alignment may not capture fine-grained relationships between languages \cite{alaux2018unsupervised}.
%Offense distance calculation error 
The coverage of offensive words in many languages is very partial, and furthermore, the ability of aligned word vectors to quantify cultural differences is limited---besides their inability to capture linguistic context, \textit{social} context (who says what to whom, where and when) is crucial to the interpretation of offensive language. This context is very seldom captured in language datasets. This limitation applies in general to the premise of offensive language detection on isolated sentences as a goal and limits the extent to which such classifiers can accurately detect offensive language in the real world.

Despite these limitations, our study serves as an important step toward understanding the role of culture in NLP and provides valuable insights for future research in this domain.

\section*{Ethics Statement}
This study involves the use of datasets with potentially harmful and offensive content, but they all follow privacy guidelines to ensure the anonymity and confidentiality of the individuals involved. The datasets used in this study are obtained from publicly available sources, and all efforts have been made to ensure compliance with relevant legal and ethical regulations. We acknowledge the potential impact of offensive language and take precautions to minimize any harm caused. Furthermore, this research aims to contribute to the development of technologies that promote a safer and more inclusive online environment. We are dedicated to fostering a culture of respect, diversity, and fairness in our research practices and encourage open dialogue on the ethical implications of offensive language detection.  

\section*{Acknowledgements}
Thanks to the anonymous reviewers for their helpful feedback. 
The authors express their gratitude to Yong Cao, Jiaang Li, and Yiyi Chen for their help during the rebuttal period and to Christoph Demus, Federico Bianchi, Aymé Arango for their assistance with the dataset collection process.
Li Zhou acknowledges financial support from China Scholarship Council (No. 202206070002).

\bibliography{anthology,custom}

\begin{thebibliography}{67}
\expandafter\ifx\csname natexlab\endcsname\relax\def\natexlab#1{#1}\fi

\bibitem[{Adelani et~al.(2022)Adelani, Neubig, Ruder, Rijhwani, Beukman, Palen-Michel, Lignos, Alabi, Muhammad, Nabende, Dione, Bukula, Mabuya, Dossou, Sibanda, Buzaaba, Mukiibi, Kalipe, Mbaye, Taylor, Kabore, Emezue, Aremu, Ogayo, Gitau, Munkoh-Buabeng, Memdjokam~Koagne, Tapo, Macucwa, Marivate, Elvis, Gwadabe, Adewumi, Ahia, Nakatumba-Nabende, Mokono, Ezeani, Chukwuneke, Oluwaseun~Adeyemi, Hacheme, Abdulmumin, Ogundepo, Yousuf, Moteu, and Klakow}]{adelani-etal-2022-masakhaner}
David Adelani, Graham Neubig, Sebastian Ruder, Shruti Rijhwani, Michael Beukman, Chester Palen-Michel, Constantine Lignos, Jesujoba Alabi, Shamsuddeen Muhammad, Peter Nabende, Cheikh M.~Bamba Dione, Andiswa Bukula, Rooweither Mabuya, Bonaventure F.~P. Dossou, Blessing Sibanda, Happy Buzaaba, Jonathan Mukiibi, Godson Kalipe, Derguene Mbaye, Amelia Taylor, Fatoumata Kabore, Chris~Chinenye Emezue, Anuoluwapo Aremu, Perez Ogayo, Catherine Gitau, Edwin Munkoh-Buabeng, Victoire Memdjokam~Koagne, Allahsera~Auguste Tapo, Tebogo Macucwa, Vukosi Marivate, Mboning~Tchiaze Elvis, Tajuddeen Gwadabe, Tosin Adewumi, Orevaoghene Ahia, Joyce Nakatumba-Nabende, Neo~Lerato Mokono, Ignatius Ezeani, Chiamaka Chukwuneke, Mofetoluwa Oluwaseun~Adeyemi, Gilles~Quentin Hacheme, Idris Abdulmumin, Odunayo Ogundepo, Oreen Yousuf, Tatiana Moteu, and Dietrich Klakow. 2022.
\newblock \href {https://aclanthology.org/2022.emnlp-main.298} {{M}asakha{NER} 2.0: {A}frica-centric transfer learning for named entity recognition}.
\newblock In \emph{Proceedings of the 2022 Conference on Empirical Methods in Natural Language Processing}, pages 4488--4508, Abu Dhabi, United Arab Emirates. Association for Computational Linguistics.

\bibitem[{Ahuja et~al.(2022)Ahuja, Kumar, Dandapat, and Choudhury}]{ahuja-etal-2022-multi}
Kabir Ahuja, Shanu Kumar, Sandipan Dandapat, and Monojit Choudhury. 2022.
\newblock \href {https://doi.org/10.18653/v1/2022.acl-long.374} {Multi task learning for zero shot performance prediction of multilingual models}.
\newblock In \emph{Proceedings of the 60th Annual Meeting of the Association for Computational Linguistics (Volume 1: Long Papers)}, pages 5454--5467, Dublin, Ireland. Association for Computational Linguistics.

\bibitem[{Alaux et~al.(2019)Alaux, Grave, Cuturi, and Joulin}]{alaux2018unsupervised}
Jean Alaux, Edouard Grave, Marco Cuturi, and Armand Joulin. 2019.
\newblock \href {https://openreview.net/forum?id=HJe62s09tX} {Unsupervised hyper-alignment for multilingual word embeddings}.
\newblock In \emph{International Conference on Learning Representations}.

\bibitem[{Arango~Monnar et~al.(2022)Arango~Monnar, Perez, Poblete, Salda{\~n}a, and Proust}]{arango-monnar-etal-2022-resources}
Ayme Arango~Monnar, Jorge Perez, Barbara Poblete, Magdalena Salda{\~n}a, and Valentina Proust. 2022.
\newblock \href {https://doi.org/10.18653/v1/2022.woah-1.12} {Resources for multilingual hate speech detection}.
\newblock In \emph{Proceedings of the Sixth Workshop on Online Abuse and Harms (WOAH)}, pages 122--130, Seattle, Washington (Hybrid). Association for Computational Linguistics.

\bibitem[{Aririguzoh(2022)}]{aririguzoh2022communication}
Stella Aririguzoh. 2022.
\newblock Communication competencies, culture and sdgs: effective processes to cross-cultural communication.
\newblock \emph{Humanities and Social Sciences Communications}, 9(1):1--11.

\bibitem[{Bassignana et~al.(2018)Bassignana, Basile, and Patti}]{bassignana2018hurtlex}
Elena Bassignana, Valerio Basile, and Viviana Patti. 2018.
\newblock \href {https://doi.org/10.4000/books.aaccademia.3085} {Hurtlex: A multilingual lexicon of words to hurt}.
\newblock In \emph{Proceedings of the Fifth Italian Conference on Computational Linguistics CLiC-it 2018: 10-12 December 2018, Torino}, Torino. Accademia University Press.

\bibitem[{Bianchi et~al.(2022)Bianchi, HIlls, Rossini, Hovy, Tromble, and Tintarev}]{bianchi-etal-2022-just}
Federico Bianchi, Stefanie HIlls, Patricia Rossini, Dirk Hovy, Rebekah Tromble, and Nava Tintarev. 2022.
\newblock \href {https://aclanthology.org/2022.emnlp-main.553} {{``}it{'}s not just hate{''}: A multi-dimensional perspective on detecting harmful speech online}.
\newblock In \emph{Proceedings of the 2022 Conference on Empirical Methods in Natural Language Processing}, pages 8093--8099, Abu Dhabi, United Arab Emirates. Association for Computational Linguistics.

\bibitem[{Blazej and Gerasimos(2021)}]{Doliki2021AnalysingTI}
Dolicki Blazej and Spanakis Gerasimos. 2021.
\newblock Analysing the impact of linguistic features on cross-lingual transfer.
\newblock \emph{ArXiv}, abs/2105.05975.

\bibitem[{Bond et~al.(1985)Bond, Wan, Leung, and Giacalone}]{bond1985responses}
Michael~H Bond, Kwok-Choi Wan, Kwok Leung, and Robert~A Giacalone. 1985.
\newblock How are responses to verbal insult related to cultural collectivism and power distance?
\newblock \emph{Journal of Cross-Cultural Psychology}, 16(1):111--127.

\bibitem[{Cao et~al.(2023)Cao, Zhou, Lee, Cabello, Chen, and Hershcovich}]{cao-etal-2023-assessing}
Yong Cao, Li~Zhou, Seolhwa Lee, Laura Cabello, Min Chen, and Daniel Hershcovich. 2023.
\newblock \href {https://aclanthology.org/2023.c3nlp-1.7} {Assessing cross-cultural alignment between {C}hat{GPT} and human societies: An empirical study}.
\newblock In \emph{Proceedings of the First Workshop on Cross-Cultural Considerations in NLP (C3NLP)}, pages 53--67, Dubrovnik, Croatia. Association for Computational Linguistics.

\bibitem[{Chen et~al.(2023)Chen, Biswas, and Bjerva}]{chen-etal-2023-colex2lang}
Yiyi Chen, Russa Biswas, and Johannes Bjerva. 2023.
\newblock \href {https://aclanthology.org/2023.nodalida-1.67} {{C}olex2{L}ang: Language embeddings from semantic typology}.
\newblock In \emph{Proceedings of the 24th Nordic Conference on Computational Linguistics (NoDaLiDa)}, pages 673--684, T{\'o}rshavn, Faroe Islands. University of Tartu Library.

\bibitem[{Collins and Kayne(2011)}]{Collins2011}
Chris Collins and Richard Kayne. 2011.
\newblock Syntactic structures of the world's languages.

\bibitem[{{\c{C}}{\"o}ltekin(2020)}]{coltekin-2020-corpus}
{\c{C}}a{\u{g}}r{\i} {\c{C}}{\"o}ltekin. 2020.
\newblock \href {https://aclanthology.org/2020.lrec-1.758} {A corpus of {T}urkish offensive language on social media}.
\newblock In \emph{Proceedings of the Twelfth Language Resources and Evaluation Conference}, pages 6174--6184, Marseille, France. European Language Resources Association.

\bibitem[{Conneau et~al.(2020{\natexlab{a}})Conneau, Khandelwal, Goyal, Chaudhary, Wenzek, Guzm{\'a}n, Grave, Ott, Zettlemoyer, and Stoyanov}]{conneau-etal-2020-unsupervised}
Alexis Conneau, Kartikay Khandelwal, Naman Goyal, Vishrav Chaudhary, Guillaume Wenzek, Francisco Guzm{\'a}n, Edouard Grave, Myle Ott, Luke Zettlemoyer, and Veselin Stoyanov. 2020{\natexlab{a}}.
\newblock \href {https://doi.org/10.18653/v1/2020.acl-main.747} {Unsupervised cross-lingual representation learning at scale}.
\newblock In \emph{Proceedings of the 58th Annual Meeting of the Association for Computational Linguistics}, pages 8440--8451, Online. Association for Computational Linguistics.

\bibitem[{Conneau et~al.(2020{\natexlab{b}})Conneau, Wu, Li, Zettlemoyer, and Stoyanov}]{conneau-etal-2020-emerging}
Alexis Conneau, Shijie Wu, Haoran Li, Luke Zettlemoyer, and Veselin Stoyanov. 2020{\natexlab{b}}.
\newblock \href {https://doi.org/10.18653/v1/2020.acl-main.536} {Emerging cross-lingual structure in pretrained language models}.
\newblock In \emph{Proceedings of the 58th Annual Meeting of the Association for Computational Linguistics}, pages 6022--6034, Online. Association for Computational Linguistics.

\bibitem[{de~Vries et~al.(2022)de~Vries, Wieling, and Nissim}]{de-vries-etal-2022-make}
Wietse de~Vries, Martijn Wieling, and Malvina Nissim. 2022.
\newblock \href {https://doi.org/10.18653/v1/2022.acl-long.529} {Make the best of cross-lingual transfer: Evidence from {POS} tagging with over 100 languages}.
\newblock In \emph{Proceedings of the 60th Annual Meeting of the Association for Computational Linguistics (Volume 1: Long Papers)}, pages 7676--7685, Dublin, Ireland. Association for Computational Linguistics.

\bibitem[{Demus et~al.(2022)Demus, Pitz, Sch{\"u}tz, Probol, Siegel, and Labudde}]{demus-etal-2022-comprehensive}
Christoph Demus, Jonas Pitz, Mina Sch{\"u}tz, Nadine Probol, Melanie Siegel, and Dirk Labudde. 2022.
\newblock \href {https://doi.org/10.18653/v1/2022.woah-1.14} {Detox: A comprehensive dataset for {G}erman offensive language and conversation analysis}.
\newblock In \emph{Proceedings of the Sixth Workshop on Online Abuse and Harms (WOAH)}, pages 143--153, Seattle, Washington (Hybrid). Association for Computational Linguistics.

\bibitem[{Deng et~al.(2022)Deng, Zhou, Sun, Zheng, Mi, Meng, and Huang}]{deng-etal-2022-cold}
Jiawen Deng, Jingyan Zhou, Hao Sun, Chujie Zheng, Fei Mi, Helen Meng, and Minlie Huang. 2022.
\newblock \href {https://aclanthology.org/2022.emnlp-main.796} {{COLD}: A benchmark for {C}hinese offensive language detection}.
\newblock In \emph{Proceedings of the 2022 Conference on Empirical Methods in Natural Language Processing}, pages 11580--11599, Abu Dhabi, United Arab Emirates. Association for Computational Linguistics.

\bibitem[{Dryer and Haspelmath(2013)}]{dryer2013world}
Matthew~S Dryer and Martin Haspelmath. 2013.
\newblock The world atlas of language structures online.

\bibitem[{Eronen et~al.(2022)Eronen, Ptaszynski, Masui, Arata, Leliwa, and Wroczynski}]{ERONEN2022}
Juuso Eronen, Michal Ptaszynski, Fumito Masui, Masaki Arata, Gniewosz Leliwa, and Michal Wroczynski. 2022.
\newblock \href {https://doi.org/https://doi.org/10.1016/j.ipm.2022.102981} {Transfer language selection for zero-shot cross-lingual abusive language detection}.
\newblock \emph{Information Processing \& Management}, 59(4):102981.

\bibitem[{Eskander et~al.(2020)Eskander, Muresan, and Collins}]{eskander-etal-2020-unsupervised}
Ramy Eskander, Smaranda Muresan, and Michael Collins. 2020.
\newblock \href {https://doi.org/10.18653/v1/2020.emnlp-main.391} {Unsupervised cross-lingual part-of-speech tagging for truly low-resource scenarios}.
\newblock In \emph{Proceedings of the 2020 Conference on Empirical Methods in Natural Language Processing (EMNLP)}, pages 4820--4831, Online. Association for Computational Linguistics.

\bibitem[{Ghosh et~al.(2021)Ghosh, Baker, Jurgens, and Prabhakaran}]{ghosh-etal-2021-detecting}
Sayan Ghosh, Dylan Baker, David Jurgens, and Vinodkumar Prabhakaran. 2021.
\newblock \href {https://doi.org/10.18653/v1/2021.wnut-1.35} {Detecting cross-geographic biases in toxicity modeling on social media}.
\newblock In \emph{Proceedings of the Seventh Workshop on Noisy User-generated Text (W-NUT 2021)}, pages 313--328, Online. Association for Computational Linguistics.

\bibitem[{Goel and Sharma(2022)}]{goel-sharma-2022-leveraging}
Divyam Goel and Raksha Sharma. 2022.
\newblock \href {https://doi.org/10.18653/v1/2022.socialnlp-1.4} {Leveraging dependency grammar for fine-grained offensive language detection using graph convolutional networks}.
\newblock In \emph{Proceedings of the Tenth International Workshop on Natural Language Processing for Social Media}, pages 45--54, Seattle, Washington. Association for Computational Linguistics.

\bibitem[{Hammarstrm et~al.(2015)Hammarstrm, Forkel, Haspelmath, and Bank}]{hammarstrm2015glottolog}
Harald Hammarstrm, Robert Forkel, Martin Haspelmath, and Sebastian Bank. 2015.
\newblock Glottolog 2.6.
\newblock \emph{Max Planck Institute for the Science of Human History}.

\bibitem[{Hershcovich et~al.(2022)Hershcovich, Frank, Lent, de~Lhoneux, Abdou, Brandl, Bugliarello, Cabello~Piqueras, Chalkidis, Cui, Fierro, Margatina, Rust, and S{\o}gaard}]{hershcovich-etal-2022-challenges}
Daniel Hershcovich, Stella Frank, Heather Lent, Miryam de~Lhoneux, Mostafa Abdou, Stephanie Brandl, Emanuele Bugliarello, Laura Cabello~Piqueras, Ilias Chalkidis, Ruixiang Cui, Constanza Fierro, Katerina Margatina, Phillip Rust, and Anders S{\o}gaard. 2022.
\newblock \href {https://doi.org/10.18653/v1/2022.acl-long.482} {Challenges and strategies in cross-cultural {NLP}}.
\newblock In \emph{Proceedings of the 60th Annual Meeting of the Association for Computational Linguistics (Volume 1: Long Papers)}, pages 6997--7013, Dublin, Ireland. Association for Computational Linguistics.

\bibitem[{Hofstede(1984)}]{hofstede1984culture}
Geert Hofstede. 1984.
\newblock \emph{Culture's consequences: International differences in work-related values}, volume~5.
\newblock sage.

\bibitem[{Jackson et~al.(2019)Jackson, Watts, Henry, List, Forkel, Mucha, Greenhill, Gray, and Lindquist}]{jackson2019emotion}
Joshua~Conrad Jackson, Joseph Watts, Teague~R Henry, Johann-Mattis List, Robert Forkel, Peter~J Mucha, Simon~J Greenhill, Russell~D Gray, and Kristen~A Lindquist. 2019.
\newblock Emotion semantics show both cultural variation and universal structure.
\newblock \emph{Science}, 366(6472):1517--1522.

\bibitem[{Jeong et~al.(2022)Jeong, Oh, Lee, Ahn, Moon, Park, and Oh}]{jeong-etal-2022-kold}
Younghoon Jeong, Juhyun Oh, Jongwon Lee, Jaimeen Ahn, Jihyung Moon, Sungjoon Park, and Alice Oh. 2022.
\newblock \href {https://aclanthology.org/2022.emnlp-main.744} {{KOLD}: {K}orean offensive language dataset}.
\newblock In \emph{Proceedings of the 2022 Conference on Empirical Methods in Natural Language Processing}, pages 10818--10833, Abu Dhabi, United Arab Emirates. Association for Computational Linguistics.

\bibitem[{Joulin et~al.(2018)Joulin, Bojanowski, Mikolov, J{\'e}gou, and Grave}]{joulin-etal-2018-loss}
Armand Joulin, Piotr Bojanowski, Tomas Mikolov, Herv{\'e} J{\'e}gou, and Edouard Grave. 2018.
\newblock \href {https://doi.org/10.18653/v1/D18-1330} {Loss in translation: Learning bilingual word mapping with a retrieval criterion}.
\newblock In \emph{Proceedings of the 2018 Conference on Empirical Methods in Natural Language Processing}, pages 2979--2984, Brussels, Belgium. Association for Computational Linguistics.

\bibitem[{Kann et~al.(2019)Kann, Cho, and Bowman}]{kann-etal-2019-towards}
Katharina Kann, Kyunghyun Cho, and Samuel~R. Bowman. 2019.
\newblock \href {https://doi.org/10.18653/v1/D19-1329} {Towards realistic practices in low-resource natural language processing: The development set}.
\newblock In \emph{Proceedings of the 2019 Conference on Empirical Methods in Natural Language Processing and the 9th International Joint Conference on Natural Language Processing (EMNLP-IJCNLP)}, pages 3342--3349, Hong Kong, China. Association for Computational Linguistics.

\bibitem[{Karamolegkou and Stymne(2021)}]{karamolegkou-stymne-2021-investigation}
Antonia Karamolegkou and Sara Stymne. 2021.
\newblock \href {https://aclanthology.org/2021.nodalida-main.32} {Investigation of transfer languages for parsing {L}atin: Italic branch vs. {H}ellenic branch}.
\newblock In \emph{Proceedings of the 23rd Nordic Conference on Computational Linguistics (NoDaLiDa)}, pages 315--320, Reykjavik, Iceland (Online). Link{\"o}ping University Electronic Press, Sweden.

\bibitem[{Ke et~al.(2017)Ke, Meng, Finley, Wang, Chen, Ma, Ye, and Liu}]{ke2017lightgbm}
Guolin Ke, Qi~Meng, Thomas Finley, Taifeng Wang, Wei Chen, Weidong Ma, Qiwei Ye, and Tie-Yan Liu. 2017.
\newblock Lightgbm: A highly efficient gradient boosting decision tree.
\newblock \emph{Advances in neural information processing systems}, 30.

\bibitem[{Khan(2014)}]{khan2014impact}
Shahid~N Khan. 2014.
\newblock Impact of hofstede's cultural dimensions on subordinate's perception of abusive supervision.
\newblock \emph{International Journal of Business and Management}, 9(12):239.

\bibitem[{Kirk et~al.(2022)Kirk, Birhane, Vidgen, and Derczynski}]{kirk-etal-2022-handling}
Hannah Kirk, Abeba Birhane, Bertie Vidgen, and Leon Derczynski. 2022.
\newblock \href {https://aclanthology.org/2022.findings-emnlp.35} {Handling and presenting harmful text in {NLP} research}.
\newblock In \emph{Findings of the Association for Computational Linguistics: EMNLP 2022}, pages 497--510, Abu Dhabi, United Arab Emirates. Association for Computational Linguistics.

\bibitem[{Lauscher et~al.(2020)Lauscher, Ravishankar, Vuli{\'c}, and Glava{\v{s}}}]{lauscher-etal-2020-zero}
Anne Lauscher, Vinit Ravishankar, Ivan Vuli{\'c}, and Goran Glava{\v{s}}. 2020.
\newblock \href {https://doi.org/10.18653/v1/2020.emnlp-main.363} {From zero to hero: {O}n the limitations of zero-shot language transfer with multilingual {T}ransformers}.
\newblock In \emph{Proceedings of the 2020 Conference on Empirical Methods in Natural Language Processing (EMNLP)}, pages 4483--4499, Online. Association for Computational Linguistics.

\bibitem[{Lin et~al.(2019)Lin, Chen, Lee, Li, Zhang, Xia, Rijhwani, He, Zhang, Ma, Anastasopoulos, Littell, and Neubig}]{lin-etal-2019-choosing}
Yu-Hsiang Lin, Chian-Yu Chen, Jean Lee, Zirui Li, Yuyan Zhang, Mengzhou Xia, Shruti Rijhwani, Junxian He, Zhisong Zhang, Xuezhe Ma, Antonios Anastasopoulos, Patrick Littell, and Graham Neubig. 2019.
\newblock \href {https://doi.org/10.18653/v1/P19-1301} {Choosing transfer languages for cross-lingual learning}.
\newblock In \emph{Proceedings of the 57th Annual Meeting of the Association for Computational Linguistics}, pages 3125--3135, Florence, Italy. Association for Computational Linguistics.

\bibitem[{Littell et~al.(2017)Littell, Mortensen, Lin, Kairis, Turner, and Levin}]{littell-etal-2017-uriel}
Patrick Littell, David~R. Mortensen, Ke~Lin, Katherine Kairis, Carlisle Turner, and Lori Levin. 2017.
\newblock \href {https://aclanthology.org/E17-2002} {{URIEL} and lang2vec: Representing languages as typological, geographical, and phylogenetic vectors}.
\newblock In \emph{Proceedings of the 15th Conference of the {E}uropean Chapter of the Association for Computational Linguistics: Volume 2, Short Papers}, pages 8--14, Valencia, Spain. Association for Computational Linguistics.

\bibitem[{Litvak et~al.(2022)Litvak, Vanetik, Liebeskind, Hmdia, and Madeghem}]{litvak-etal-2022-offensive}
Marina Litvak, Natalia Vanetik, Chaya Liebeskind, Omar Hmdia, and Rizek~Abu Madeghem. 2022.
\newblock \href {https://aclanthology.org/2022.lrec-1.396} {Offensive language detection in {H}ebrew: can other languages help?}
\newblock In \emph{Proceedings of the Thirteenth Language Resources and Evaluation Conference}, pages 3715--3723, Marseille, France. European Language Resources Association.

\bibitem[{Liu et~al.(2022)Liu, Kong, Huang, Mao, and Xue}]{liu-etal-2022-multiple}
Jiexi Liu, Dehan Kong, Longtao Huang, Dinghui Mao, and Hui Xue. 2022.
\newblock \href {https://aclanthology.org/2022.findings-emnlp.546} {Multiple instance learning for offensive language detection}.
\newblock In \emph{Findings of the Association for Computational Linguistics: EMNLP 2022}, pages 7387--7396, Abu Dhabi, United Arab Emirates. Association for Computational Linguistics.

\bibitem[{Lwowski et~al.(2022)Lwowski, Rad, and Rios}]{lwowski-etal-2022-measuring}
Brandon Lwowski, Paul Rad, and Anthony Rios. 2022.
\newblock \href {https://aclanthology.org/2022.coling-1.574} {Measuring geographic performance disparities of offensive language classifiers}.
\newblock In \emph{Proceedings of the 29th International Conference on Computational Linguistics}, pages 6600--6616, Gyeongju, Republic of Korea. International Committee on Computational Linguistics.

\bibitem[{Ma et~al.(2022)Ma, Datta, Wang, and Vosoughi}]{ma-etal-2022-encbp}
Weicheng Ma, Samiha Datta, Lili Wang, and Soroush Vosoughi. 2022.
\newblock \href {https://doi.org/10.18653/v1/2022.findings-acl.221} {{E}n{CBP}: A new benchmark dataset for finer-grained cultural background prediction in {E}nglish}.
\newblock In \emph{Findings of the Association for Computational Linguistics: ACL 2022}, pages 2811--2823, Dublin, Ireland. Association for Computational Linguistics.

\bibitem[{Malaviya et~al.(2017)Malaviya, Neubig, and Littell}]{malaviya-etal-2017-learning}
Chaitanya Malaviya, Graham Neubig, and Patrick Littell. 2017.
\newblock \href {https://doi.org/10.18653/v1/D17-1268} {Learning language representations for typology prediction}.
\newblock In \emph{Proceedings of the 2017 Conference on Empirical Methods in Natural Language Processing}, pages 2529--2535, Copenhagen, Denmark. Association for Computational Linguistics.

\bibitem[{Mandl et~al.(2019)Mandl, Modha, Majumder, Patel, Dave, Mandlia, and Patel}]{mandl2019overview}
Thomas Mandl, Sandip Modha, Prasenjit Majumder, Daksh Patel, Mohana Dave, Chintak Mandlia, and Aditya Patel. 2019.
\newblock Overview of the hasoc track at fire 2019: Hate speech and offensive content identification in indo-european languages.
\newblock In \emph{Proceedings of the 11th forum for information retrieval evaluation}, pages 14--17.

\bibitem[{McGillivray et~al.(2022)McGillivray, Alahapperuma, Cook, Di~Bonaventura, Mero{\~n}o-Pe{\~n}uela, Tyson, and Wilson}]{mcgillivray-etal-2022-leveraging}
Barbara McGillivray, Malithi Alahapperuma, Jonathan Cook, Chiara Di~Bonaventura, Albert Mero{\~n}o-Pe{\~n}uela, Gareth Tyson, and Steven Wilson. 2022.
\newblock \href {https://aclanthology.org/2022.evonlp-1.7} {Leveraging time-dependent lexical features for offensive language detection}.
\newblock In \emph{Proceedings of the The First Workshop on Ever Evolving NLP (EvoNLP)}, pages 39--54, Abu Dhabi, United Arab Emirates (Hybrid). Association for Computational Linguistics.

\bibitem[{Moran et~al.(2014)Moran, McCloy, and Wright}]{moran2014phoible}
Steven Moran, Daniel McCloy, and Richard Wright. 2014.
\newblock Phoible online.

\bibitem[{Mubarak et~al.(2021)Mubarak, Rashed, Darwish, Samih, and Abdelali}]{mubarak-etal-2021-arabic}
Hamdy Mubarak, Ammar Rashed, Kareem Darwish, Younes Samih, and Ahmed Abdelali. 2021.
\newblock \href {https://aclanthology.org/2021.wanlp-1.13} {{A}rabic offensive language on {T}witter: Analysis and experiments}.
\newblock In \emph{Proceedings of the Sixth Arabic Natural Language Processing Workshop}, pages 126--135, Kyiv, Ukraine (Virtual). Association for Computational Linguistics.

\bibitem[{Nozza(2021)}]{nozza-2021-exposing}
Debora Nozza. 2021.
\newblock \href {https://doi.org/10.18653/v1/2021.acl-short.114} {Exposing the limits of zero-shot cross-lingual hate speech detection}.
\newblock In \emph{Proceedings of the 59th Annual Meeting of the Association for Computational Linguistics and the 11th International Joint Conference on Natural Language Processing (Volume 2: Short Papers)}, pages 907--914, Online. Association for Computational Linguistics.

\bibitem[{Papadakis et~al.(2022)Papadakis, Aletta, Kang, Oberman, Mitchell, and Stavroulakis}]{PAPADAKIS2022109031}
Nikolaos~M. Papadakis, Francesco Aletta, Jian Kang, Tin Oberman, Andrew Mitchell, and Georgios~E. Stavroulakis. 2022.
\newblock \href {https://doi.org/https://doi.org/10.1016/j.apacoust.2022.109031} {Translation and cross-cultural adaptation methodology for soundscape attributes – a study with independent translation groups from english to greek}.
\newblock \emph{Applied Acoustics}, 200:109031.

\bibitem[{Patankar et~al.(2022)Patankar, Gokhale, Litake, Mandke, and Kadam}]{patankar-etal-2022-train}
Shantanu Patankar, Omkar Gokhale, Onkar Litake, Aditya Mandke, and Dipali Kadam. 2022.
\newblock \href {https://aclanthology.org/2022.sumeval-1.2} {To train or not to train: Predicting the performance of massively multilingual models}.
\newblock In \emph{Proceedings of the First Workshop on Scaling Up Multilingual Evaluation}, pages 8--12, Online. Association for Computational Linguistics.

\bibitem[{Patil et~al.(2022)Patil, Talukdar, and Sarawagi}]{patil-etal-2022-overlap}
Vaidehi Patil, Partha Talukdar, and Sunita Sarawagi. 2022.
\newblock \href {https://doi.org/10.18653/v1/2022.acl-long.18} {Overlap-based vocabulary generation improves cross-lingual transfer among related languages}.
\newblock In \emph{Proceedings of the 60th Annual Meeting of the Association for Computational Linguistics (Volume 1: Long Papers)}, pages 219--233, Dublin, Ireland. Association for Computational Linguistics.

\bibitem[{Pires et~al.(2019)Pires, Schlinger, and Garrette}]{pires-etal-2019-multilingual}
Telmo Pires, Eva Schlinger, and Dan Garrette. 2019.
\newblock \href {https://doi.org/10.18653/v1/P19-1493} {How multilingual is multilingual {BERT}?}
\newblock In \emph{Proceedings of the 57th Annual Meeting of the Association for Computational Linguistics}, pages 4996--5001, Florence, Italy. Association for Computational Linguistics.

\bibitem[{Plank and Agi{\'c}(2018)}]{plank-agic-2018-distant}
Barbara Plank and {\v{Z}}eljko Agi{\'c}. 2018.
\newblock \href {https://doi.org/10.18653/v1/D18-1061} {Distant supervision from disparate sources for low-resource part-of-speech tagging}.
\newblock In \emph{Proceedings of the 2018 Conference on Empirical Methods in Natural Language Processing}, pages 614--620, Brussels, Belgium. Association for Computational Linguistics.

\bibitem[{Ptaszynski et~al.(2019)Ptaszynski, Pieciukiewicz, and Dybała}]{publ152265}
Michal Ptaszynski, Agata Pieciukiewicz, and Paweł Dybała. 2019.
\newblock \href {http://2019.poleval.pl/files/poleval2019.pdf} {Results of the poleval 2019 shared task 6 : first dataset and open shared task for automatic cyberbullying detection in polish twitter}.
\newblock In Maciej Ogrodniczuk and Łukasz Kobyliński, editors, \emph{Proceedings of the PolEval 2019 Workshop}, pages 89--110. Institute of Computer Sciences. Polish Academy of Sciences, Warszawa.

\bibitem[{Rahimi et~al.(2019)Rahimi, Li, and Cohn}]{rahimi-etal-2019-massively}
Afshin Rahimi, Yuan Li, and Trevor Cohn. 2019.
\newblock \href {https://doi.org/10.18653/v1/P19-1015} {Massively multilingual transfer for {NER}}.
\newblock In \emph{Proceedings of the 57th Annual Meeting of the Association for Computational Linguistics}, pages 151--164, Florence, Italy. Association for Computational Linguistics.

\bibitem[{Santy et~al.(2023)Santy, Liang, Le~Bras, Reinecke, and Sap}]{santyliang2023nlpositionality}
Sebastin Santy, Jenny~T. Liang, Ronan Le~Bras, Katharina Reinecke, and Maarten Sap. 2023.
\newblock Nlpositionality: Characterizing design biases of datasets and models.
\newblock In \emph{Annual Meeting of the Association for Computational Linguistics (ACL)}.

\bibitem[{Sigurbergsson and Derczynski(2020)}]{sigurbergsson-derczynski-2020-offensive}
Gudbjartur~Ingi Sigurbergsson and Leon Derczynski. 2020.
\newblock \href {https://aclanthology.org/2020.lrec-1.430} {Offensive language and hate speech detection for {D}anish}.
\newblock In \emph{Proceedings of the Twelfth Language Resources and Evaluation Conference}, pages 3498--3508, Marseille, France. European Language Resources Association.

\bibitem[{Sittar and Mladenic(2023)}]{sittar2023classification}
Abdul Sittar and Dunja Mladenic. 2023.
\newblock Classification of cross-cultural news events.
\newblock \emph{arXiv preprint arXiv:2301.05543}.

\bibitem[{S{\o}gaard(2022)}]{sogaard-2022-ban}
Anders S{\o}gaard. 2022.
\newblock \href {https://aclanthology.org/2022.emnlp-main.351} {Should we ban {E}nglish {NLP} for a year?}
\newblock In \emph{Proceedings of the 2022 Conference on Empirical Methods in Natural Language Processing}, pages 5254--5260, Abu Dhabi, United Arab Emirates. Association for Computational Linguistics.

\bibitem[{Srinivasan et~al.(2022)Srinivasan, Kholkar, Kejriwal, Ganu, Dandapat, Sitaram, Santhanam, Aditya, Bali, and Choudhury}]{Srinivasan_2022}
Anirudh Srinivasan, Gauri Kholkar, Rahul Kejriwal, Tanuja Ganu, Sandipan Dandapat, Sunayana Sitaram, Balakrishnan Santhanam, Somak Aditya, Kalika Bali, and Monojit Choudhury. 2022.
\newblock \href {https://doi.org/10.1609/aaai.v36i11.21736} {Litmus predictor: An ai assistant for building reliable, high-performing and fair multilingual nlp systems}.
\newblock \emph{Proceedings of the AAAI Conference on Artificial Intelligence}, 36(11):13227--13229.

\bibitem[{Srinivasan et~al.(2021)Srinivasan, Sitaram, Ganu, Dandapat, Bali, and Choudhury}]{srinivasan2021predicting}
Anirudh Srinivasan, Sunayana Sitaram, Tanuja Ganu, Sandipan Dandapat, Kalika Bali, and Monojit Choudhury. 2021.
\newblock \href {http://arxiv.org/abs/2110.08875} {Predicting the performance of multilingual nlp models}.

\bibitem[{Sun et~al.(2021)Sun, Ahn, Park, Tsvetkov, and Mortensen}]{sun-etal-2021-cross}
Jimin Sun, Hwijeen Ahn, Chan~Young Park, Yulia Tsvetkov, and David~R. Mortensen. 2021.
\newblock \href {https://doi.org/10.18653/v1/2021.eacl-main.204} {Cross-cultural similarity features for cross-lingual transfer learning of pragmatically motivated tasks}.
\newblock In \emph{Proceedings of the 16th Conference of the European Chapter of the Association for Computational Linguistics: Main Volume}, pages 2403--2414, Online. Association for Computational Linguistics.

\bibitem[{Taras et~al.(2016)Taras, Steel, and Kirkman}]{taras2016does}
Vas Taras, Piers Steel, and Bradley~L Kirkman. 2016.
\newblock Does country equate with culture? beyond geography in the search for cultural boundaries.
\newblock \emph{Management International Review}, 56:455--487.

\bibitem[{Tran and Bisazza(2019)}]{tran-bisazza-2019-zero}
Ke~Tran and Arianna Bisazza. 2019.
\newblock \href {https://doi.org/10.18653/v1/D19-6132} {Zero-shot dependency parsing with pre-trained multilingual sentence representations}.
\newblock In \emph{Proceedings of the 2nd Workshop on Deep Learning Approaches for Low-Resource NLP (DeepLo 2019)}, pages 281--288, Hong Kong, China. Association for Computational Linguistics.

\bibitem[{Van~Oudenhoven et~al.(2008)Van~Oudenhoven, de~Raad, Askevis-Leherpeux, Boski, Brunborg, Carmona, Barelds, Hill, Mla{\v{c}}i{\'c}, Motti et~al.}]{van2008terms}
Jan~Pieter Van~Oudenhoven, Boele de~Raad, Francoise Askevis-Leherpeux, Pawel Boski, Geir~Scott Brunborg, Carmen Carmona, Dick Barelds, Charles~T Hill, Boris Mla{\v{c}}i{\'c}, Frosso Motti, et~al. 2008.
\newblock Terms of abuse as expression and reinforcement of cultures.
\newblock \emph{International Journal of Intercultural Relations}, 32(2):174--185.

\bibitem[{Wu and Dredze(2019)}]{wu-dredze-2019-beto}
Shijie Wu and Mark Dredze. 2019.
\newblock \href {https://doi.org/10.18653/v1/D19-1077} {Beto, bentz, becas: The surprising cross-lingual effectiveness of {BERT}}.
\newblock In \emph{Proceedings of the 2019 Conference on Empirical Methods in Natural Language Processing and the 9th International Joint Conference on Natural Language Processing (EMNLP-IJCNLP)}, pages 833--844, Hong Kong, China. Association for Computational Linguistics.

\bibitem[{Zampieri et~al.(2019)Zampieri, Malmasi, Nakov, Rosenthal, Farra, and Kumar}]{zampieri-etal-2019-predicting}
Marcos Zampieri, Shervin Malmasi, Preslav Nakov, Sara Rosenthal, Noura Farra, and Ritesh Kumar. 2019.
\newblock \href {https://doi.org/10.18653/v1/N19-1144} {Predicting the type and target of offensive posts in social media}.
\newblock In \emph{Proceedings of the 2019 Conference of the North {A}merican Chapter of the Association for Computational Linguistics: Human Language Technologies, Volume 1 (Long and Short Papers)}, pages 1415--1420, Minneapolis, Minnesota. Association for Computational Linguistics.

\bibitem[{Zhou et~al.(2023)Zhou, Cabello, Cao, and Hershcovich}]{zhou-etal-2023-cross}
Li~Zhou, Laura Cabello, Yong Cao, and Daniel Hershcovich. 2023.
\newblock \href {https://aclanthology.org/2023.c3nlp-1.2} {Cross-cultural transfer learning for {C}hinese offensive language detection}.
\newblock In \emph{Proceedings of the First Workshop on Cross-Cultural Considerations in NLP (C3NLP)}, pages 8--15, Dubrovnik, Croatia. Association for Computational Linguistics.

\end{thebibliography}
\bibliographystyle{acl_natbib}

\appendix
\section{\textsc{HarmCheck} List}
\label{app:harmcheck}

\begin{itemize}
    \item \textbf{Risk of harm protocol}:  The specific risks of harm are derived from offensive language detection datasets in multiple languages. Experimenters for this work have access to these datasets, but they don't need to read them carefully.
    \item \textbf{Preview}: The work includes a Content Warning directly after Abstract. The content warning is in italics to maximize visibility.  
    \item \textbf{Distance}: Only a few offensive words are included to explain cross-cultural differences in the introduction section.
    \item \textbf{Respect}: For slurs and profanities in text, the  authors star out the first vowel with an asterisk (“*”).
\end{itemize}

\section{Data-dependent and Language-dependent Features}
\label{app:D-L-features}
The existing features used to predict transfer learning are mainly divided into data-dependent and language-dependent features.

\subsection{Data-dependent Features}
Data-dependent features primarily manifest in three aspects: dataset size, Type-Token Ratio (TTR), and word overlap, which are specific measurements obtained from the datasets themselves.

\paragraph{Dataset size.} 
% Data-specific features are statistical features of the particular dataset used, which cannot represent or distinguish a specific language.
\textit{Transfer Size} and \textit{Target Size}, serving as fundamental features, indicate the number of data entries utilized for training in transfer and target datasets, denoted by $\mathcal{S}_\mathrm{1}$ and $\mathcal{S}_\mathrm{2}$ respectively. Additionally, the \textit{Ratio Size} $\mathcal{S}_r = \frac{\mathcal{S}_\mathrm{1}}{\mathcal{S}_\mathrm{2}}$ is utilized to indicate the relative size difference between the transfer dataset and the target dataset. It quantifies how much larger the transfer dataset is in comparison to the target dataset.

\paragraph{Type-Token Ratio} TTR is the ratio between the number of types (vocabulary set) and the number of tokens in a dataset, measuring lexical variation. A higher TTR represents a higher lexical variation. 
We utilize \textit{Transfer TTR} $\mathrm{TTR}_1$ and \textit{Target TTR} $\mathrm{TTR}_2$ to represent the Type-Token Ratio (TTR) of the transfer and target datasets, respectively. The lexical-variation \textit{Distance TTR} between the transfer and target datasets is defined as $\mathrm{TTR}_d = \left( 1 - \frac{\mathrm{TTR}_1}{\mathrm{TTR}_2} \right) ^2$.

\paragraph{Word Overlap} \textit{Word overlap} is used to measure the lexical similarity between a pair of languages (datasets). It is defined as $\frac{|V_1\cap V_2|}{|V_1|+|V_2|}$, where $V_1$ and $V_2$ represent the vocabularies of transfer dataset and target dataset.

% To explore the transfer prediction ability of different dimension feature spaces for OLD, inspired by~\citet{sun-etal-2021-cross}, we first adopt five feature groups: Data-specific, Typology, Geography, Orthography, Pragmatic. Moreover, to explore the impact of cultural differences, we introduce a new Cultural group.

\subsection{Language-dependent Features}
Language-dependent features encompass six linguistic distances queried from the URIEL Typological Database~\cite{littell-etal-2017-uriel}, as well as three linguistic features (PRAG) that manifest in linguistic patterns and quantify distinct aspects of language pragmatics~\cite{sun-etal-2021-cross}.

\paragraph{URIEL.} The feature vector in URIEL encompasses various linguistic features that describe the typological properties of languages, which include \textit{Genetic}, \textit{Syntactic}, \textit{Phonological}, \textit{Inventory}, \textit{Geographic} and \textit{Featural}. The derived distances quantify the similarities or dissimilarities between languages based on these features.

% \paragraph{Typology} Typology feature includes five distance features evaluated for language pairs across different dimensional spaces from the URIEL Typological Database~\cite{littell-etal-2017-uriel}.

\begin{itemize}
    \item \textit{Genetic:} The Genetic distance derived from the Glottolog tree~\cite{hammarstrm2015glottolog} of language families, where it quantifies the dissimilarity between two languages by measuring their distance within the tree.
    \item \textit{Syntactic:} The syntactic distance is the cosine distance between syntax features of language pairs. The syntax features are adapted from the World Atlas of Language Structures (WALS)~\cite{dryer2013world}, Syntactic Structures of World Languages (SSWL)~\cite{Collins2011} and Ethnologue \footnote{\url{https://www.ethnologue.com/}}.
    \item \textit{Phonological:} The phonological distance measure the cosine distance between vectors containing phonological information from Ethnologue and WALS.
    \item \textit{Inventory:} The inventory distance is the cosine distance between the inventory feature vectors of languages, sourced from the PHOIBLE database~\cite{moran2014phoible}.
    \item \textit{Geographical:} Geographical distance can represent the shortest distance between two languages on the surface of Earth’s sphere. It is another component of URIEL, in which the geographical vectors of each language express geographical location with a fixed number of dimensions and each dimension representing the same feature. 
    \item \textit{Featural:} The cosine distance between vectors incorporating features from the above five.
\end{itemize}

% \paragraph{Geography} 

\paragraph{PRAG.} These features are pragmatically-inspired linguistic features that capture cross-cultural similarities manifested in linguistic patterns.

% Pragmatic features quantify the contextual factors in language use, which adopt Type-Token Ratio~\cite{richards1987type} to measure lexical diversity and three kinds of pragmatically-motivated features~\cite{sun-etal-2021-cross}.

\begin{figure*}[ht]
    \centering
    \includegraphics[width=0.95\linewidth]{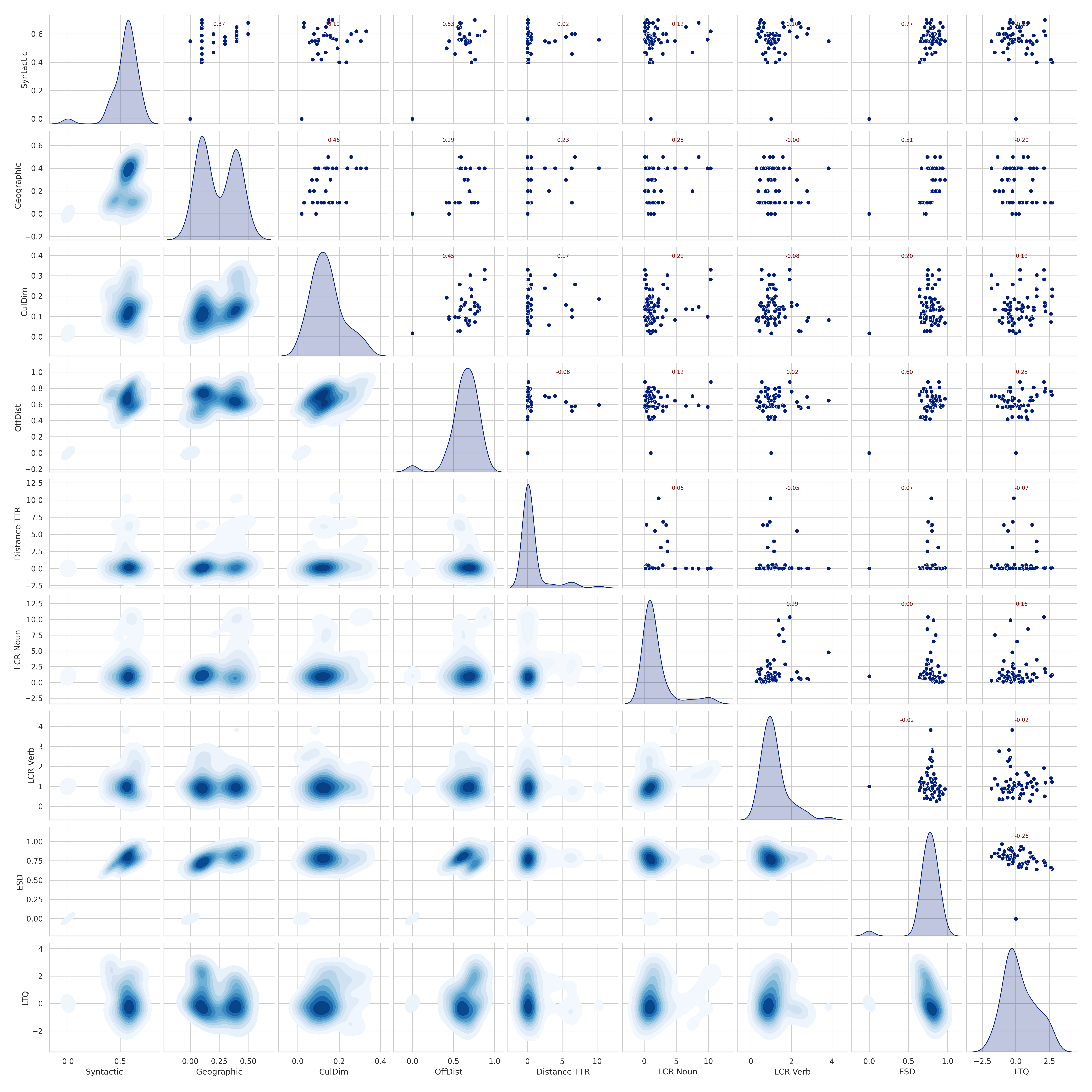}
    \caption{A pairplot showcasing the relationship between culturally relevant features using kernel density estimation (KDE), accompanied by correlation coefficients (in red digit) annotated on the corresponding scatter plots. The diagonal plots display the distribution of each feature using KDE curves, while the lower triangle depicts the joint distributions and contour plots using a color gradient (Blues colormap).}
    \label{fig:pairplot}
\end{figure*}

\begin{itemize}
    \item \textit{Language Context-level Ratio} (\textit{LCR}) measures the extent to which the language leaves the identity of entities and predicates to context. It only takes into account noun probabilities and verb probabilities here, which represent the likelihood of nouns and verbs occurring in the text, respectively.
    \item \textit{Literal Translation Quality} (\textit{LTQ}) quantifies how well a given language pair's MWEs are preserved in literal (word-by-word) translation, using a bilingual dictionary.
    \item \textit{Emotion Semantics Distance} (\textit{EDS}) measures how similarly emotions are lexicalized across languages.
\end{itemize}

\section{Feature correlation analysis}
\label{app:correlation}
\subsection{Different distances}
In this section, we discuss the correlations between our different features, with a specific focus on data-independent features. Among all the features, we have selected those that are associated with cultural diversity, such as \textit{CulDim}, \textit{Geographical}, \textit{Syntactic}, and so on. We construct a sample set by sampling from the OLD zero-shot experiment, which involves all possible pairs of data for the relevant features. Figure~\ref{fig:pairplot} shows the relationship between the selected features.
As depicted in Figure~\ref{fig:pairplot}, we observe a strong positive correlation between \textit{CulDim} and \textit{Geographical}, suggesting that there is a relationship between cultural differences and geographical differences to some extent. The correlations between different features vary significantly, highlighting the importance of introducing multidimensional features to a certain extent.

\subsection{Cultural dimension}
\begin{figure}[ht]
    \centering
    \includegraphics[width=1\linewidth]{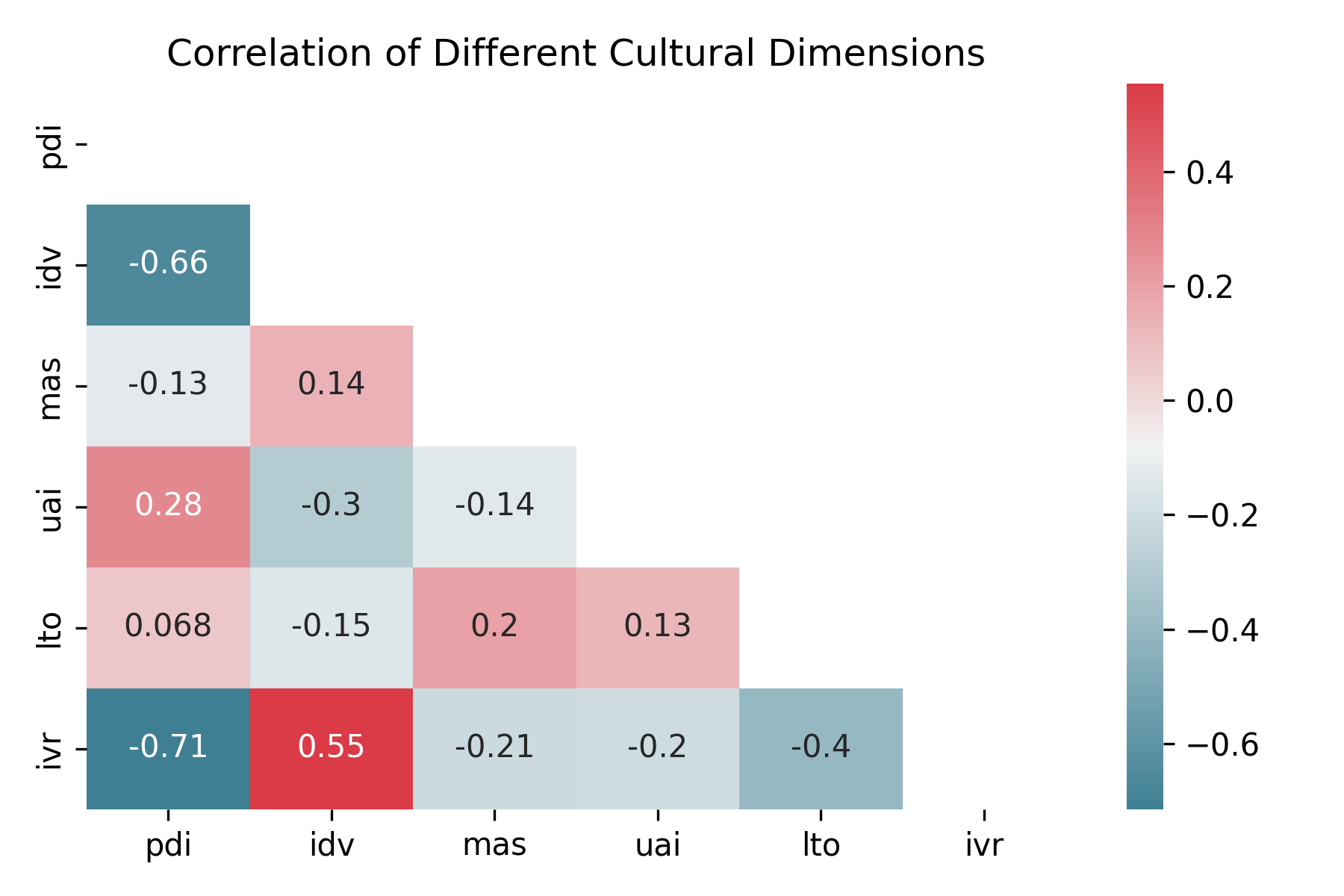}
    \caption{Fine-grained correlation analysis of dimension features in \textit{CulDim}.}
    \label{fig:Cor_cul}
\end{figure}

% \citet{bond1985responses}
% \citet{khan2014impact}
We conduct a fine-grained correlation analysis for each dimension feature in the \textit{CulDim}. As shown in the Figure~\ref{fig:Cor_cul}, \textit{Indulgence (ivr)} and \textit{Individualism (idv)} exhibit a relatively strong positive correlation, implying that societies or groups that tend to score high on indulgence also tend to score high on individualism. This could indicate a cultural inclination towards personal freedom, self-expression, and a focus on individual rights and achievements. 
However, \textit{Power Distance (pdi)} and has strong negative correlation with \textit{Indulgence (ivr)} and\textit{Individualism (idv)}.
This indicates that societies or groups characterized by a high power distance, where hierarchical structures and authority are emphasized, tend to have lower levels of indulgence and individualism. It suggests a cultural tendency towards obedience, respect for authority, and collective orientation over personal freedom and self-expression.

\section{Datasets details}
% \subsection{Datasets}
\label{app:Datasets}
% \textbf{ArabicOLD}~\cite{mubarak-etal-2021-arabic} is an Arabic offensive language tweet dataset that is not biased by specific dialects, topics, or genres, in which each tweet is labeled as clean or offensive, including special tags for vulgar tweets and hate speech. The annotation guidelines are developed jointly with an experienced annotator, who is a native Arabic speaker and has a good knowledge of various Arabic dialects.

\textbf{COLD}~\cite{deng-etal-2022-cold} is a Chinese dataset, covering the topics of racial, gender, and regional bias. It is collected by strategies of keyword querying and related sub-topics crawling, and labeld by 17 native Chinese native workers.

\textbf{ChileOLD}~\cite{arango-monnar-etal-2022-resources} contains tweets in the Spanish language that originated in Chile, which is a representative of the Spanish spoken in South America. Each tweet is annotated with several fine-grained offensive categories by three native Chileans.

\textbf{DeTox}~\cite{demus-etal-2022-comprehensive} is a German dataset that is far more comprehensive with twelve different annotation categories. The annotation schema is developed with first-hand users from a reporting office for offensive comments in Germany. The dataset is collected by manually creating keyword list with the help of Google Trends.

\textbf{HinHD}~\cite{mandl2019overview}, utilized in the HASOC 2019 shared task, was collected from Twitter using hashtags and keywords associated with offensive content. In the labeling process, multiple junior annotators participated via an online system to evaluate the tweets.

\textbf{KOLD}~\cite{jeong-etal-2022-kold} is a Korean offensive language dataset comprising comments from NAVER news and YouTube platform, which is also filtered by using predefined keywords.

\textbf{NJH}~\cite{bianchi-etal-2022-just} is a English dataset collected with keywords and associated with immigration in the US and/or UK. It is annotated by ten undergraduate researchers from universities in the UK and US, based on which the dataset can be split into \textit{NJH-UK} and \textit{NJH-US}.

\textbf{PolEval}~\cite{publ152265} is the first dataset for the Polish language containing annotations of harmful and toxic language. 
The dataset was automatically collected from Twitter accounts in the Polish language and annotated by layperson volunteers under the supervision of a cyberbullying and hate-speech expert.

\textbf{TurkishOLD}~\cite{coltekin-2020-corpus} is the first corpus of offensive language for Turkish and consists of randomly sampled micro-blog posts from Twitter. The annotators are native speakers of Turkish, and all are highly educated.

% \textbf{SOLD}~\cite{ranasinghe2022sold}

% \textbf{DKHATE}~\cite{sigurbergsson-derczynski-2020-offensive}

% \textbf{OGTD}~\cite{pitenis-etal-2020-offensive}

\section{Table Summary for Supplementary Information}
 By presenting this information in a table format, it allows for easy reference and understanding of the selection criteria and categorization of the data. This section serves to enhance the clarity and comprehensiveness of the paper, ensuring that readers have access to all the necessary details regarding the countries, languages, and their corresponding abbreviations used in the study.

\subsection{ Overview of Countries, Languages, and Groups in Feature Analysis}
\label{app:CL}
To provide a concise overview the countries and languages included in the feature analysis, we provides a detailed Table~\ref{tab:lang_coun} showcasing  the abbreviations used to represent each country and language in the feature analysis figures (Figure~\ref{fig:analysis_cultural} and Figure~\ref{fig:analysis_off}), as well as their corresponding groups based on relevant criteria.

% \usepackage{multirow}
\begin{table}[ht]
\centering
\scalebox{0.8}{
\begin{tabular}{l|ll|ll}
\toprule
                                           & \multicolumn{2}{c|}{\textbf{Country}}        & \multicolumn{2}{c}{\textbf{Language}} \\ \midrule
{\multirow{3}{*}{\textbf{East Asia}}}      & CN & {China}              & zho             & Chinese             \\
                                           & JP & {Japan}              & jpn             & Japanese            \\
                                           & KR & {South Korea}        & kor             & Korean              \\ \midrule
{\multirow{2}{*}{\textbf{South Asia}}}     & IN & {India}              & hin             & Hindi               \\
                                           &    & {}                   & tam             & Tamil               \\ \midrule
{\multirow{5}{*}{\textbf{Western Europe}}} & NL & {Netherlands}        & nld             & Dutch               \\
                                           & FR & {France}             & fra             & France              \\
                                           & UK & {\multirow{2}{2cm}{the United Kingdom}} & eng             & English             \\\\
                                           & ES & {Spain}              & spa             & Spanish             \\
                                           & DE & {German}             & deu             & German              \\ \midrule
{\multirow{3}{*}{\textbf{East Europe}}}    & PL & {Poland}             & pol             & Polish              \\
                                           & RU & {Russia}             & rus             & Russian             \\
                                           &    &                      & ces             & Czech               \\ \midrule
{\multirow{3}{*}{\textbf{Middle East}}}    & TR & {Turkey}             & tur             & Turkish             \\
                                           & JO & {Jordan}             & ara             & Arabic              \\
                                           & IR & {Iran}               & fas             & Persian             \\ \midrule
{\textbf{South America}}                   & CL & {Chile}              &                 &                     \\ \cline{1-3}
{\textbf{North America}}                   & US & {\multirow{2}{2cm}{the United States}}  &                 &                     \\\\ \bottomrule

\end{tabular}}
\caption{Overview of Countries, Languages, and Groups in Feature Analysis}
\label{tab:lang_coun}
\end{table}

\subsection{Cultural Dimension Values of Countries and Languages}
\label{app:Cultural_value}
We display the cultural dimension values associated with the countries and languages involved in our experiments.
In OLD, the cultural dimension values are depicted in Figure~\ref{tab:cultural}. The datasets used in OLD contain country information, allowing for the direct utilization of corresponding cultural dimension values. 
However, in the case of SA and DEP tasks, where only language information is provided in the datasets, direct indexing of cultural dimension values is not feasible. As a solution, we select countries where each language is considered the sole official language and employ the average cultural dimension values of these countries to represent the cultural features associated with the respective language. The value details are shown in Table~\ref{tab:all-value}.

\begin{table}[ht]
\centering
\scalebox{0.85}{
\begin{tabular}{l|rrrrrr}
\toprule
\textbf{Country}  & \textbf{pdi} & \textbf{idv} & \textbf{mas} & \textbf{uai} & \textbf{lto} & \textbf{ivr} \\
\midrule
China  & 80  & 20  & 66  & 30  & 87  & 24  \\
Chile  & 63   & 23  & 28  & 86  & 31  & 68  \\
Germany  & 35   & 67  & 66  & 65  & 83  & 40  \\
India  & 77  & 48  & 56  & 40  & 51  & 26  \\
South Korea  &  60  & 18  & 39  & 85  & 100  & 29  \\
United Kingdom  & 35   & 89  & 66  & 35  & 51  & 69  \\
United States  &  40  & 91  & 62  & 46  & 26  & 68  \\
Poland  &  68  & 60  & 64  & 93  & 38  & 29  \\
Turkey  &  66  & 37  & 45  & 85  & 46  & 49  \\
\bottomrule
\end{tabular}
}
\caption{Cultural-value features in OLD.}
\label{tab:cultural}
\end{table}

% \usepackage{multirow}
\begin{table*}[ht]
\centering
\scalebox{0.85}{
\begin{tabular}{llrrrrrr}
\toprule
\multirow{2}{*}{\textbf{Language}} & \multirow{2}{*}{\textbf{Country}} & \multicolumn{6}{c}{\textbf{Cultural Value}}                                                                                                                                                                     \\ \cline{3-8} 
                                   &                                   & \multicolumn{1}{l}{\textbf{pdi}} & \multicolumn{1}{l}{\textbf{idv}} & \multicolumn{1}{l}{\textbf{mas}} & \multicolumn{1}{l}{\textbf{uai}} & \multicolumn{1}{l}{\textbf{lto}} & \multicolumn{1}{l}{\textbf{ivr}} \\ \hline
\midrule
\multirow{10}{*}{Arabic}           & Egypt                             & 80                               & 37                               & 55                               & 55                               & 42                               & 0                                \\
                                   & Jordan                            & 70                               & 30                               & 45                               & 65                               & 16                               & 43                               \\
                                   & Kuwait                            & 90                               & 25                               & 40                               & 80                               & -1                               & -1                               \\
                                   & Lebanon                           & 62                               & 43                               & 48                               & 57                               & 22                               & 10                               \\
                                   & Libya                             & 100                              & 35                               & 66                               & 67                               & 15                               & 74                               \\
                                   & Qatar                             & 93                               & 25                               & 55                               & 80                               & -1                             & -1                             \\
                                   & Saudi Arabia                      & 72                               & 48                               & 43                               & 64                               & 27                               & 14                               \\
                                   & Syria                             & 80                               & 35                               & 52                               & 60                               & 30                               & -1                               \\
                                   & Tunisia                           & 70                               & 40                               & 40                               & 75                               & -1                             & -1                             \\
                                   & United Arab Emirates              & 74                               & 36                               & 52                               & 66                               & 22                               & 22                               \\
\midrule
Chinese                            & China                             & 80                               & 20                               & 66                               & 30                               & 87                               & 24                               \\
\midrule
Czech                              & Czech Republic                    & 57                               & 58                               & 57                               & 74                               & 70                               & 29                               \\
\midrule
\multirow{2}{*}{Dutch}             & Netherlands                       & 38                               & 80                               & 14                               & 53                               & 67                               & 68                               \\
                                   & Suriname                          & 85                               & 47                               & 37                               & 92                               & -1                               & -1                               \\
\midrule
\multirow{10}{*}{English}          & Australia                         & 38                               & 90                               & 61                               & 51                               & 21                               & 71                               \\
                                   & Ghana                             & 80                               & 15                               & 40                               & 65                               & 4                                & 72                               \\
                                   & Jamaica                           & 45                               & 39                               & 68                               & 13                               & -1                               & -1                               \\
                                   & Namibia                           & 65                               & 30                               & 40                               & 45                               & 35                               & -1                               \\
                                   & Nigeria                           & 80                               & 30                               & 60                               & 55                               & 13                               & 84                               \\
                                   & Sierra Leone                      & 70                               & 20                               & 40                               & 50                               & -1                               & -1                               \\
                                   & Trinidad and Tobago               & 47                               & 16                               & 58                               & 55                               & 13                               & 80                               \\
                                   & United Kingdom                    & 35                               & 89                               & 66                               & 35                               & 51                               & 69                               \\
                                   & United States                     & 40                               & 91                               & 62                               & 46                               & 26                               & 68                               \\
                                   & Zambia                            & 60                               & 35                               & 40                               & 50                               & 30                               & 42                               \\
\midrule
\multirow{3}{*}{France}            & Burkina Faso                      & 70                               & 15                               & 50                               & 55                               & 27                               & 18                               \\
                                   & France                            & 68                               & 71                               & 43                               & 86                               & 63                               & 48                               \\
                                   & Senegal                           & 70                               & 25                               & 45                               & 55                               & 25                               & -1                               \\
\midrule
\multirow{2}{*}{German}            & Austria                           & 11                               & 55                               & 79                               & 70                               & 60                               & 63                               \\
                                   & Germany                           & 35                               & 67                               & 66                               & 65                               & 83                               & 40                               \\
\midrule
Hindi                              & India                             & 77                               & 48                               & 56                               & 40                               & 51                               & 26                               \\
\midrule
Japanese                           & Japan                             & 54                               & 46                               & 95                               & 92                               & 88                               & 42                               \\
\midrule
Korean                             & South Korea                       & 60                               & 18                               & 39                               & 85                               & 100                              & 29                               \\
\midrule
Persian                            & Iran                              & 58                               & 41                               & 43                               & 59                               & 14                               & 40                               \\
\midrule
Polish                             & Poland                            & 68                               & 60                               & 64                               & 93                               & 38                               & 29                               \\
\midrule
Russian                            & Russia                            & 93                               & 39                               & 36                               & 95                               & 81                               & 20                               \\
\midrule
\multirow{12}{*}{Spanish}          & Argentina                         & 49                               & 46                               & 56                               & 86                               & 20                               & 62                               \\
                                   & Chile                             & 63                               & 23                               & 28                               & 86                               & 31                               & 68                               \\
                                   & Colombia                          & 67                               & 13                               & 64                               & 80                               & 13                               & 83                               \\
                                   & Costa Rica                        & 35                               & 15                               & 21                               & 86                               & -1                               & -1                               \\
                                   & Dominican Republic                & 65                               & 30                               & 65                               & 45                               & 13                               & 54                               \\
                                   & El Salvador                       & 66                               & 19                               & 40                               & 94                               & 20                               & 89                               \\
                                   & Guatemala                         & 95                               & 6                                & 37                               & 98                               & -1                               & -1                               \\
                                   & Honduras                          & 80                               & 20                               & 40                               & 50                               & -1                               & -1                               \\
                                   & Mexico                            & 81                               & 30                               & 69                               & 82                               & 24                               & 97                               \\
                                   & Panama                            & 95                               & 11                               & 44                               & 86                               & -1                               & -1                               \\
                                   & Spain                             & 57                               & 51                               & 42                               & 86                               & 48                               & 44                               \\
                                   & Uruguay                           & 61                               & 36                               & 38                               & 98                               & 26                               & 53                               \\
\midrule
Tamil                              & Sri Lanka                         & 80                               & 35                               & 10                               & 45                               & 45                               & -1                               \\
\midrule
Turkish                            & Turkey                            & 66                               & 37                               & 45                               & 85                               & 46                               & 49                               \\ 
\bottomrule
\end{tabular}}
\caption{Cultural Dimension Values of Languages in SA and DEP Tasks. In particular,
When the dimension score is $-1$, it indicates that the country does not have a score in that specific cultural dimension.}
\label{tab:all-value}
\end{table*}

\subsection{Feature List of Baselines and Feature Groups}
\label{app:feature_group}
Table~\ref{tab:feature} provides a comprehensive list of features utilized in the baselines and feature groups employed for comparison in the study. It showcases specific sets together for analysis and evaluation purposes.

% \usepackage{multirow}
\begin{sidewaystable*}[ht]
\resizebox{\linewidth}{!}{
\begin{tabular}{l|l|cc|ccccc|c|cc}
\hline
\multirow{2}{*}{\textbf{Type}}     & \multirow{2}{*}{\textbf{Feature}} & \multicolumn{2}{c|}{\textbf{Baselines}} & \multicolumn{5}{c|}{\textbf{Feature Group}}                                                                 & \textbf{Ours}           & \multicolumn{2}{c}{\textbf{Other Group}} \\ \cline{3-12} 
                                   &                                   & \textbf{LangRank}    & \textbf{MTVEC}   & \textbf{Data-specific} & \textbf{Topology} & \textbf{Geography} & \textbf{Orthography} & \textbf{Pragmatic} & \textbf{Cultural-value}    & \textbf{PRAG}       \\ \hline
\multirow{7}{*}{Data-dependent}    & Transfer Size           & $\sqrt{}$                    &                  & $\sqrt{}$                      &                   &                    &                      &                    &                            &                     \\
                                   & Target Size                  & $\sqrt{}$                    &                  & $\sqrt{}$                      &                   &                    &                      &                    &                            &                     \\
                                   & Ratio Size              & $\sqrt{}$                    &                  & $\sqrt{}$                      &                   &                    &                      &                    &                              &                     \\
                                   & Transfer TTR                     & $\sqrt{}$                    &                  &                        &                   &                    &                      & $\sqrt{}$                  &                              &                     \\
                                   & Target TTR                         & $\sqrt{}$                    &                  &                        &                   &                    &                      & $\sqrt{}$                  &                                   &                     \\
                                   & Distance TTR                     & $\sqrt{}$                    &                  &                        &                   &                    &                      & $\sqrt{}$                  &                                 &                     \\
                                   & Word overlap                     & $\sqrt{}$                    &                  &                        &                   &                    & $\sqrt{}$                    &                    &                         &                     \\ \hline
\multirow{17}{*}{Data-independent} & Genetic                           & $\sqrt{}$                    &                  &                        & $\sqrt{}$                 &                    &                      &                    &                            &                     \\
                                   & Syntactic                         & $\sqrt{}$                    &                  &                        & $\sqrt{}$                 &                    &                      &                    &                          &                     \\
                                   & Featural                          & $\sqrt{}$                    &                  &                        & $\sqrt{}$                 &                    &                      &                    &                                 &                     \\
                                   & Phonological                      & $\sqrt{}$                    &                  &                        & $\sqrt{}$                 &                    &                      &                    &                                 &                     \\
                                   & Inventory                         & $\sqrt{}$                    &                  &                        &                   & $\sqrt{}$                  &                      &                    &                                      &                     \\
                                   & Geographic                        & $\sqrt{}$                    &                  &                        &                   &                    &                      &                    &                                   &                     \\
                                   & $\text{LCR}_{verb}$ Distance                    &                      &                  &                        &                   &                    &                      & $\sqrt{}$                  &                                  & $\sqrt{}$                   \\
                                   & $\text{LCR}_{noun}$ Distance                    &                      &                  &                        &                   &                    &                      & $\sqrt{}$                  &                                     & $\sqrt{}$                   \\
                                   & ESD Distance                     &                      &                  &                        &                   &                    &                      & $\sqrt{}$                  &                          & $\sqrt{}$                   \\
                                   & LTQ score                        &                      &                  &                        &                   &                    &                      & $\sqrt{}$                  &                            & $\sqrt{}$                   \\
                                   & Rep\_Diff                         &                      & $\sqrt{}$                &                        &                   &                    &                      &                    &                                &                     \\
                                   & Power Distance (pdi)                               &                      &                  &                        &                   &                    &                      &                    & $\sqrt{}$                       &                     \\
                                   & Individualism (idv)                               &                      &                  &                        &                   &                    &                      &                    & $\sqrt{}$                       &                     \\
                                   & Masculinity (mas)                               &                      &                  &                        &                   &                    &                      &                    & $\sqrt{}$                         &                     \\
                                   & Uncertainty Avoidance (uai)                               &                      &                  &                        &                   &                    &                      &                    & $\sqrt{}$                  &                     \\
                                   & Long-Term Orientation (lto)                               &                      &                  &                        &                   &                    &                      &                    & $\sqrt{}$                &                     \\
                                   & Indulgence (ivr)                               &                      &                  &                        &                   &                    &                      &                    & $\sqrt{}$                         &                     \\ \hline
\end{tabular}}
\caption{Feature Groups and Baseline Features in the Study}
\label{tab:feature}
\end{sidewaystable*}

\end{document}